\documentclass{ieeeaccess}
\usepackage{cite}
\usepackage{amsmath,amssymb,amsfonts}
\usepackage{graphicx}
\usepackage{textcomp}
\usepackage{booktabs}
\usepackage{listings}
\usepackage{xcolor}
\let\accessyear\year
\def\year{2026}
\usepackage{tikz}
\usetikzlibrary{arrows.meta, positioning, shapes.geometric}
\let\year\accessyear
\year{2026}
\definecolor{accessblue}{cmyk}{1,0.3,0,0.2}
\makeatletter
\long\def\@makecaption#1#2{%
  \ifx\@captype\@IEEEtablestring
    \begin{flushleft}
    \vspace*{5pt}
    {\vss\color{accessblue}\tablecapheadfont #1. \ }{\raggedright\tablecapfont#2\vss}%
    \end{flushleft}
    \@IEEEtablecaptionsepspace
  \else
    \@IEEEfigurecaptionsepspace
    {\raggedright\noindent{\color{accessblue}\figcapheadfont #1. \ }\figcapfont#2\par}%
  \fi}
\makeatother
\makeatletter\input{t1ptm.fd}\input{t1phv.fd}\input{t1pcr.fd}\makeatother
\DeclareFontShape{T1}{ptm}{n}{n}{<->ssub*ptm/m/n}{}
\DeclareFontShape{T1}{ptm}{n}{it}{<->ssub*ptm/m/it}{}
\DeclareFontShape{T1}{ptm}{n}{sl}{<->ssub*ptm/m/sl}{}
\DeclareFontShape{T1}{ptm}{n}{sc}{<->ssub*ptm/m/sc}{}
\DeclareFontShape{T1}{phv}{n}{n}{<->ssub*phv/m/n}{}
\DeclareFontShape{T1}{phv}{n}{it}{<->ssub*phv/m/it}{}
\DeclareFontShape{T1}{pcr}{n}{n}{<->ssub*pcr/m/n}{}
\makeatletter
\def\ps@titlepage{%
  \def\@oddhead{\parbox[t]{\textwidth}{\mbox{}\\[-6.3mm]\mbox{}\nheaderfont\color{accessblue}\headName\hfill\mbox{}\color{black}\vspace*{-1.5mm}\par\hrulefill}}
  \def\@evenhead{\parbox[t]{\textwidth}{\mbox{}\\[-6.3mm]\mbox{}\headerlogo\hfill\nheaderfont\color{accessblue}\headName{\raisebox{-2pt}{\,}}\color{black}\vspace*{-1.5mm}\par\hrulefill}}
  \def\@oddfoot{\raisebox{3pt}{%
    \hbox to 0pc{\hbox to \textwidth{{\footervolfont VOLUME\ \thevol, \theyear}\hfill\mbox{}}}%
    \hbox to 0pc{\hbox to \textwidth{\mbox{}\hfill{\footervolfont\@pubid}\hfill\mbox{}}}%
    \hbox to 0pc{\hbox to \textwidth{\mbox{}\hfill{\footerpagefont\thepage}}}}}%
  \def\@evenfoot{\raisebox{3pt}{%
    \hbox to 0pc{\hbox to \textwidth{{\footerpagefont\thepage}\hfill}\mbox{}}\hspace*{1.8pc}%
    \hspace*{-1.8pc}\hbox to 0pc{\hbox to \textwidth{\mbox{}\hfill{\footervolfont\@pubid}\hfill\mbox{}}}%
    \hbox to 0pc{\hbox to \textwidth{\mbox{}\hfill{\footervolfont VOLUME\ \thevol, \theyear}}}\hspace*{1.8pc}}}}
\def\ps@headings{%
  \def\@oddhead{\parbox[t]{\textwidth}{\mbox{}\\[-6.3mm]{\raisebox{-2pt}{\headerfont\rightmark}}\hfill\mbox{}\vspace*{-1.5mm}\par\hrulefill}}
  \def\@evenhead{\parbox[t]{\textwidth}{\mbox{}\\[-6.3mm]\mbox{}\headerlogoall\hfill{\raisebox{-2pt}{\headerfont\leftmark}}\vspace*{-1.5mm}\par\hrulefill}}
  \def\@oddfoot{\raisebox{8pt}{%
    \hbox to 0pc{\hbox to \textwidth{{\footervolfont VOLUME\ \thevol, \theyear}\hfill\mbox{}}}%
    \hbox to 0pc{\hbox to \textwidth{\mbox{}\hfill{\footerpagefont\thepage}}}}}%
  \def\@evenfoot{\raisebox{8pt}{%
    \hbox to 0pc{\hbox to \textwidth{{\footerpagefont\thepage}\mbox{}\hfill}}\hspace*{1.8pc}%
    \hspace*{-1.8pc}\hbox to 0pc{\hbox to \textwidth{\mbox{}\hfill{\footervolfont VOLUME\ \thevol, \theyear}}}\hspace*{1.8pc}}}}
\ps@headings
\makeatother
\usepackage{etoolbox}
\makeatletter
\patchcmd{\appendices}{\def\thesectiondis{APPENDIX \Alph{section}}.}{\def\thesectiondis{APPENDIX \Alph{section}}}{}{\ClassWarning{ieeeaccess}{appendices patch failed}}
\makeatother
\makeatletter

\newif\if@biographyTOCentrynotmade \@biographyTOCentrynotmadetrue
\makeatother
\usepackage[hyphens]{url}
\usepackage[hidelinks,bookmarks=false]{hyperref}
\begin{document}
\history{Date of publication xxxx 00, 0000, date of current version xxxx 00, 0000.}
\doi{10.1109/ACCESS.2026.DOI}

\title{Declarative Skills for AI Agents in Knowledge-Grounded Tool-Use Workflows}
\author{\uppercase{M. Danish Lim}\authorrefmark{1},
\uppercase{I. Danial Bin Sharudin}\authorrefmark{1},
\uppercase{Wen Han Chen}\authorrefmark{1},
\uppercase{Cedric Lim}\authorrefmark{1},
and \uppercase{Laura Wynter}\authorrefmark{1}}
\address[1]{School of Computing and Information Systems, Singapore Management University, 80 Stamford Road, Singapore 178902 (e-mail: muhammadlim.2022@scis.smu.edu.sg; danials.2022@scis.smu.edu.sg; wenhan.chen.2022@scis.smu.edu.sg; cedric.lim.2022@scis.smu.edu.sg; lwynter@smu.edu.sg)}
\tfootnote{This work was supported by the Singapore Ministry of Education (MOE) Academic Research Fund (AcRF) Tier 1 Grant under Proposal ID 24-SIS-SMU-074.}

\markboth
{Lim \headeretal: Declarative Skills for AI Agents in Knowledge-Grounded Tool-Use Workflows}
{Lim \headeretal: Declarative Skills for AI Agents in Knowledge-Grounded Tool-Use Workflows}

\corresp{Corresponding author: Laura Wynter (e-mail: lwynter@smu.edu.sg).}

\begin{abstract}
We study how orchestration affects tool-using large language model (LLM) agents in
customer-service workflows over an unstructured knowledge base, comparing a declarative agent that
reads natural-language skill files, an imperative agent built on a state machine, and an
unscaffolded baseline. The three are formalized as policy classes within one decentralized
partially observable Markov decision process and evaluated on a complete design of five models,
two retrieval regimes, and three agents, with paired tests. Averaged over all conditions,
orchestration does not improve task success. The effect, however, depends on the quality of the
agent's observation channel. Under perfect retrieval, skill files are neutral-to-negative on four
of the five models; under embedding retrieval they are neutral-to-positive on four of the five. On the tasks where baseline
and scaffold disagree, the declarative win rate rises from 42.4\% to 62.7\% (p = 0.016), and the
same interaction is present for the state machine (p = 0.003). Orchestration thus substitutes for
information the agent lacks rather than supplementing information it has.
Since deployed systems never attain perfect retrieval, the regime in which skill files help is the
realistic one. Skill files also improve write-argument accuracy in
all ten conditions (p = 0.002) without improving end-to-end success, and reduce unauthorized
writes only where a compliance gap exists. The state machine lowers task success on four of the five models, and its verification gate is
phase routing that misclassification can bypass.
Enforcing the same precondition at the tool-execution layer eliminates unauthorized writes while
preserving task success.
\end{abstract}


\titlepgskip=-15pt

\maketitle

\section{Introduction}

\PARstart{T}{ool-using} artificial intelligence (AI) agents extend generative language models with the ability to act on external systems, and evaluating how well they perform real-world tasks has become a central research problem as such agents are deployed at scale. We focus on a general and practically important class of agent: customer-service workflows, which combine three distinct competencies: \emph{conversational} (greeting the user, eliciting and clarifying intent), \emph{procedural} (determining ordering constraints, verifications required, and tool-call sequences), and \emph{reasoning} (interpreting retrieved documents to choose the correct tool arguments). These competencies are broadly applicable beyond banking customer service.

We make use of $\tau$-Knowledge~\cite{tauknowledge}, an extension of
$\tau$-Bench~\cite{taubench}, which was developed for evaluating AI agents on tasks that require 
coordinated retrieval over a large unstructured knowledge base (KB). Unlike prior agent benchmarks that supply a
fully-specified tool interface up front,
$\tau$-Knowledge allows the capabilities to be discoverable; for example, many state-changing
operations are referenced only in natural-language documentation and must be
located via knowledge-base  search before they can be  invoked.
This is framed as a decentralised partially observable Markov decision process,  (Dec-POMDP)~\cite{decpomdp} over a shared state space
$S = S_{\text{DB}} \times S_{\text{history}}$, where $S_\text{DB}$ is the state of the database which includes user and 
system entities, and 
$S_\text{history}$ is the stored user-AI agent conversation  history.
The AI agent observes
only tool outputs and user messages, and the binary task reward depends on whether the
final database state $S_\text{DB}$  matches a hand-curated target. The problem is a POMDP because the AI agent and the user 
have asymmetric and incomplete views of the state, $S$.

We propose two paradigms to orchestrate the AI agents. Imperative orchestration is defined as  externalised, deterministic control. Deterministic control has broad appeal in combating the non-determinism of large language models (LLMs). It can be argued that end-to-end
LLM inference for real-world tasks is too risky  and that some of
the control should be externalised to deterministic code. Recursive Language
Models (RLMs)~\cite{rlm2025} are an aggressive form of this idea
 that proposes to treat LLM input context as an external
environment (e.g.\ a Python read--eval--print loop (REPL) variable). Then, using RLM, the LLM agent programmatically
inspects and decomposes the REPL environment, and recursively invokes
itself on smaller sub-problems, composing the final
answer from its sub-results. The same general idea is found also in LangGraph deterministic finite automata (DFAs)~\cite{langgraph}, ReAct reasoning-action loops ~\cite{yao2023react}, and numerous other works. In our instantiation, our ImperativeAgent owns the phase graph and the
transition rules (such as verification before write, bounded retries) while the LLM is invoked as a per-phase sub-routine. The desired benefit of the ImperativeAgent
is that this deterministic enforcement will reduce hallucination, improve
interpretability, and make compliance properties more auditable.

The second form of  orchestration  we examine is called declarative. This paradigm relies on an agent skills approach similar to that  proposed by Anthropic\cite{agentskills}, who suggest that  procedural knowledge should be expressed in natural language and read by the model
at run time  as needed. An agent skill is a markdown document describing when an action is
appropriate, the preconditions and ordering constraints, and the
 tool-argument requirements. The LLM interprets the agent skill files
 as part of its system prompt and chooses the workflow and details
on-line. The justification for the declarative paradigm is that  the LLM's own attention mechanism can integrate natural language skill content with
retrieved  evidence in more flexible ways that a fixed state graph. 

In summary, we define the following two orchestrated agents:
\begin{itemize}
    \item  ImperativeAgent that implements a finite-state machine with deterministic transitions, explicit verification gates, and hard-coded retry policies, representing externalised control and programmatic orchestration.
    \item   DeclarativeAgent that follows Anthropic-style Agent Skills: the model reads a small set of markdown skill files in its system prompt and is free to choose the workflow, tools, and verification strategy in natural language, with no explicit state machine.
\end{itemize}

We are interested in the following question. For tool-using AI agents in realistic, knowledge-grounded customer-service workflows, how does natural-language declarative orchestration (skill files) compare with programmatic imperative orchestration (a state machine) and with an unscaffolded baseline, on task success, compliance, and cost? We thus propose  three research questions, stated formally in Section~\ref{sec:formulation}: whether declarative skill files improve task success over the baseline (RQ1); whether an imperative verification gate reduces unauthorized writes (RQ2); and whether retrieval quality moderates any orchestration advantage (RQ3). Both declarative and imeprative paradigms have a priori plausible arguments. The imperative approach promises to reduce hallucination through deterministic enforcement, and the declarative approach should  be less brittle. We test both against the unscaffolded LLM agent used in the $\tau$-Knowledge benchmark~\cite{tauknowledge}.

Casting all three agents as policy classes over a shared Dec-POMDP allows us to isolate
orchestration style as the independent variable, and evaluating with paired significance tests,
absent from much of the agent-orchestration literature, lets us separate real effects from
sampling noise. Our central result concerns the conditions under which orchestration helps.

Averaged over the full design, neither scaffold improves task success: the declarative agent's
apparent Pass$^1$ gains do not survive paired testing, and the imperative state machine
significantly lowers success on four of the five models. This average is of limited use, however,
because it aggregates over an interaction. The value of orchestration depends on the quality of
the agent's observation channel. Under perfect retrieval, where the agent is handed exactly the
policy text a task requires, skill files are neutral-to-negative on four of five models. Under
embedding retrieval, where the retrieved evidence is noisy or incomplete, they are
neutral-to-positive on four of the five. Pooling every \emph{discordant} task across models, meaning
every task that the baseline and the scaffold do not both pass or both fail, which are the only
tasks a paired test draws information from, the declarative win rate against the baseline rises
from 42.4\% to 62.7\% ($\chi^2=5.78$, $p=0.016$); the same
interaction holds independently for the state machine ($p=0.003$), which stays net-harmful in both
regimes but markedly less so when retrieval degrades.

We interpret this as orchestration substituting for information the agent lacks rather than
supplementing information it already has. A skill file encodes procedural knowledge about how a
task should be carried out. When retrieval already supplies that knowledge, restating it adds
little and the additional instruction may distract; when retrieval fails, the skill file is the
only source of that knowledge. Perfect retrieval is an idealisation, used here as an upper bound,
that no deployed system attains, so the condition under which skill files are useful is the one
deployments operate in.

Three further results are consistent with this interpretation. First, skill files improve
write-argument accuracy in all ten model$\times$retrieval conditions (sign test $p=0.002$) while moving
end-to-end Pass$^1$ not at all, which is what one expects when an intervention improves individual
actions on tasks that require a median of eight state-changing writes to all be correct. Second,
skill files reduce unauthorized writes only on the models that have a compliance gap to close, and
do nothing on the models that are already compliant, which is the same substitution logic on a
different axis. Third, and most actionable for regulated deployments, the imperative agent's
``verification gate'' turns out to be phase-routing over an inferred control state that
misclassification can bypass; re-binding the identical precondition to the write tools themselves
eliminates unauthorized writes entirely (0 of 467) while preserving task success. Guarantees belong
on the action, not on the agent's belief about which phase it is in.

The paper proceeds as follows. In the next section, we discuss related work. Then, Section~\ref{sec:formulation} casts the three agents as policy classes within a Dec-POMDP and states our three main research questions. Sections~\ref{sec:declarative}--\ref{sec:imperative} describe the DeclarativeAgent and ImperativeAgent in detail. Section~\ref{sec:theory} presents a conceptual framework relating the three policy classes to a set of testable hypotheses. Section~\ref{sec:results} provides the experimental design and the main  results and Section~\ref{sec:safety} provides an ablation into the compliance and efficiency of the agents. We conclude with a discussion that ties our  findings  back to our research questions. The appendix includes additional details.

\section{Related Work}

The application domain used in  $\tau$-Knowledge is $\tau$-Banking,   a
fintech-customer-service set of tasks. The benchmark environment contains
698 documents ($\sim$195K tokens, 71 topics, 21 product categories),
14 permanent agent tools plus {51 discoverable tools,
and 97 evaluation tasks. Each task requires, on average,
18.6 documents and 9.52 tool calls (max 33) to resolve. The benchmark scoring proposed in  \cite{tauknowledge}  is  $\text{pass}^k$, and while the paper considers $k=1,3$, we focus only on the more challenging pass$^1$ metric. Documents provided include product specifications,
internal procedural policies (e.g.\ retention protocols, account-closure
eligibility), and discoverable-tool signatures with required-argument
schemas. Tool names contain random
    four-digit suffixes (e.g.\ \texttt{close\_\allowbreak{}bank\_\allowbreak{}account\_\allowbreak{}7392})
    that cannot be guessed.
    The benchmark distinguishes two ways the agent obtains task-critical documents. In the \emph{gold} setting the minimal required document set is placed directly in the system prompt; in the \emph{retrieval} setting the agent must find documents through an external retriever, which the benchmark instantiates with keyword matching (BM25), dense embeddings, and an interactive \emph{terminal} variant in which the agent issues its own search commands. For clarity we use a single pair of terms throughout this paper: \emph{golden} for the gold setting and \emph{embedding} for our dense-retrieval configuration; where we quote the benchmark's published numbers (Table~\ref{tab:tauk-baselines}) we retain its column names, including \emph{Terminal}.

Beyond $\tau$-Knowledge, several recent works benchmark customer-support agents on adherence to business policies, 
multi-step workflows, and tool-use correctness. 
\cite{businessadherence,tauextension}. 
These works assume a fixed orchestration style, comprising a single LLMAgent-like loop with tools, and focus on model or retriever comparisons. Our
 work proposes to explore the benefits of agent skills via our DeclarativeAgent as well as programmatic approaches to invoking tools via our Imperative Agent under a common benchmark and tool set.

Our work also relates to the literature on agentic scaffolding and orchestration. 
ReAct-style reasoning-and-acting loops~\cite{yao2023react} interleave natural-language thoughts and tool calls. 
Graph-based and DFA-style frameworks such as LangGraph~\cite{langgraph} expose the agent's control flow as an explicit state machine or  graph, 
allowing deterministic transitions. 
Recursive Language Models~\cite{rlm2025} externalise control further, treating the prompt as an environment variable, and allow the LLM to recursively call itself over decomposed subproblems, demonstrating strong gains on long-context reasoning tasks~\cite{rlmblog}. Our ImperativeAgent is in this broad family of deterministic approaches to orchestration. A parallel line of work leaves control flow to the model but adds a learned or verbalised feedback loop, as in Reflexion's episodic self-critique~\cite{shinn2023reflexion}; our DeclarativeAgent differs in supplying procedural knowledge \emph{ahead} of execution rather than correction after it.

A parallel line of work studies retrieval-augmented generation in realistic, messy knowledge bases. 
$\tau$-Knowledge itself highlights that even frontier models struggle to retrieve, interpret, and act on unstructured documentation~\cite{tauknowledge}, 
and subsequent reports have underscored how retrieval quality often dominates model choice in customer-support agents~\cite{tauknowledge,sierra2026bench}. There is also increasing interest in replacing brittle tool registries and Model Context Protocol (MCP)-style plugins with file-centric agent interfaces, 
where files serve simultaneously as context, tools, and skills~\cite{filesareallyouneed}.  
Our results strengthen these findings: golden retrieval exposes the benefits of skill-file orchestration, while noisy embedding retrieval sharply reduces performance for all agents, 
illustrating that agent skills and high-capacity LLMs with reasoning cannot compensate for fundamentally incorrect evidence.

Related to our declarative orchestration is Anthropic's Agent Skills specification~\cite{agentskills}, 
which proposes reusable SKILL.md files as composable, model-readable procedural knowledge for agents. Agent Skills are loaded via progressive disclosure, with short metadata always in context and full skill bodies read only when needed~\cite{agentskills,agent-skills-blog}. 
Follow-on work has generalised this idea to  other ecosystems, 
arguing that skills should be small, focused markdown files that can be swapped or combined to tailor agent behaviour~\cite{filesareallyouneed,agent-skills-blog}. 

Our DeclarativeAgent instantiates this paradigm, using three skill files to encode conversational structure, banking procedures, and knowledge-discovery strategy, 
and provides, to our knowledge, the first systematic comparison between a skill-file declarative agent and a programmatic state-machine agent on a realistic customer-support benchmark.

The authors of \cite{tauknowledge} identified the main causes of failure when using their benchmark on LLM agents as: (1) \emph{complex interdependencies between offerings}
($\sim$14.5\% of failures): multi-hop reasoning across documents to find
the optimal product combination; (2) \emph{failure to respect implicit
subtask ordering} ($\sim$5\%), e.g.\ disputes must resolve before credit
limit increases; (3) \emph{overtrusting user assertions} ($\sim$4\%):
acting on user-claimed state without verifying via tools; and
(4) \emph{search inefficiency and unwarranted assumptions} ($\sim$23\%):
committing to early hypotheses rather than searching the KB. 

These causes of failure motivate our ImperativeAgent and DeclarativeAgent strategies. 
While failure type 1 may be assumed to be tied mainly to LLM capacity (and hence model parameter count), we aim to rectify failures 2--3, namely topological task ordering
and verification gating using code with our ImperativeAgent. 
Similarly, we provide explicit KB-search guidance in the agent skills of our DeclarativeAgent, 
positioning declarative skills as a low-cost capability enhancement for AI agents.

The
$\tau$-Knowledge paper evaluates five frontier models across the various retrieval
configurations. Their main finding is that the benchmark is
hard for current LLM agents: their best non-gold configuration was GPT-5.2 (high reasoning)
with terminal use at 25.52\% $\text{pass}^1$, and even with gold
documents provided to the agent in context their best score was
Claude-4.5-Opus (high) at 39.69\%. Our Table~\ref{tab:tauk-baselines} reproduces
their benchmark's main $\text{pass}^1$ results using their unscaffolded LLM agent~\cite[Table~2]{tauknowledge}.

\begin{table*}[!t]
\centering
\caption{$\tau$-Knowledge benchmark's frontier-model baselines on unscaffolded LLM agents using $\text{pass}^1$ (\%), reproduced from~\cite{tauknowledge} \emph{for context}. These are the benchmark authors' published results on models we do not otherwise evaluate; our own experiments (Section~\ref{sec:results}) use four different, lighter-weight models and are not directly comparable to this table. ``Gold'' provides the minimal document set to the agent in context; ``Terminal'' is the benchmark's interactive setting in which the agent issues its own retrieval commands in a terminal. Parentheses indicate $\Delta$ vs.\ the Gold setting for each row; ``Reas.'' is the reasoning-level setting. Bold marks the best value in each column.}
\label{tab:tauk-baselines}
\small
\begin{tabular}{llrrrrr}
\toprule
                  &           &        & \multicolumn{2}{c}{Embeddings} &         &           \\
\cmidrule(lr){4-5}
Model             & Reas. & Gold   & text-emb-3-large & Qwen3-emb-8B & BM25    & Terminal  \\
\midrule
GPT-5.2           & High      & 32.73  & \textbf{23.45} ($-9.3$)   & \textbf{24.74} ($-8.0$) & \textbf{24.48} ($-8.2$) & \textbf{25.52} ($-7.2$) \\
GPT-5.2           & None      & 15.72  & \phantom{0}8.25 ($-7.5$) & 12.37 ($-3.4$) & \phantom{0}9.54 ($-6.2$) & 11.60 ($-4.1$) \\
Claude-4.5-Op   & High      & \textbf{39.69}  & 18.30 ($-21.4$) & 19.59 ($-20.1$) & 17.78 ($-21.9$) & 24.74 ($-14.9$) \\
Claude-4.5-Son & High      & 33.76  & 17.53 ($-16.2$) & 17.78 ($-16.0$) & 16.75 ($-17.0$) & 22.42 ($-11.3$) \\
Gemini-3-Pro      & High      & 33.25  & 12.89 ($-20.4$) & 12.89 ($-20.4$) & 13.66 ($-19.6$) & 15.72 ($-17.5$) \\
Gemini-3-Flash    & High      & 36.34  & 18.56 ($-17.8$) & 18.56 ($-17.8$) & 18.56 ($-17.8$) & 20.62 ($-15.7$) \\
\bottomrule
\end{tabular}
\end{table*}

\section{Problem Formulation}
\label{sec:formulation}

We model the customer-service interaction as a finite-horizon, two-agent decentralised partially-observable Markov decision process (Dec-POMDP)~\cite{decpomdp}.
A simulation, hereafter referred to as a \emph{task}, is one rollout of this process on a fixed task specification drawn from $\tau$-Knowledge.

The world state is $S = S_{\text{DB}} \times S_{\text{conv}}$, where $S_{\text{DB}}$ is the relational state of the banking database (customer records, accounts, transactions, disputes, cards) and $S_{\text{conv}}$ is the rolling conversation history. Two policies operate over $S$ jointly: the task agent $\pi$ that we design, and a user-simulator $\pi_u$ whose persona and intent are fixed by the task. Both observe $S$ only through messages and tool outputs; the database is not directly visible to either agent.

At each turn $t$, the task agent emits an action $a_t \in \mathcal{A}$ given an information state $h_t = (o_{1:t}, a_{1:t-1})$. We partition the action space as $\mathcal{A} = \mathcal{A}_{\text{say}} \cup \mathcal{A}_{\text{read}} \cup \mathcal{A}_{\text{write}}$:
$\mathcal{A}_{\text{say}}$ are natural-language turns directed at the user; $\mathcal{A}_{\text{read}}$ are non-mutating tool calls (\texttt{KB\_\allowbreak{}search}, account lookups, tool discovery, identity \texttt{log\_\allowbreak{}verification}); $\mathcal{A}_{\text{write}}$ are state-mutating tool calls that change $S_{\text{DB}}$ (transaction submissions, account changes, referrals, transfers to human agents, etc.).
A task terminates when either party emits a stop token, when an unrecoverable error is raised, or when a horizon $T_{\max}$ is reached. Reward is binary,
\[
r(\tau) \;=\; 1\!\left\{ S_{\text{DB}}^{\text{final}} = S_{\text{DB}}^{\text{gold}}\right\} \cdot 1\!\left\{ A^{\text{required}} \subseteq A(\tau) \right\},
\]
where $A^{\text{required}}$ is the set of gold action-checks (canonical write tools with their canonical arguments) supplied by the benchmark and $A(\tau)$ is the multiset of tool calls actually issued. Across $K$ trials, the standard $\tau$-Knowledge metric is $\mathrm{pass}^k = \mathbb{E}_{\tau_{1:K}}[1\{\forall i \in [k],\, r(\tau_i)=1\}]$; we report $\mathrm{pass}^1$ as our primary metric.

We compare three policy classes within this Dec-POMDP. Let $\theta_{\text{sys}}$ denote the baseline system prompt distributed with the benchmark, and let the LLM under evaluation be $M$.

\textbf{Definition 1 (Baseline policy).} The baseline policy is
$\pi_B(a_t \mid h_t;\, \theta_{\text{sys}}, M) = M\!\left(a_t \mid \theta_{\text{sys}}, h_t\right)$,
i.e.\ the model conditions only on the static system prompt and the running history. There is no agent-side control flow.

\textbf{Definition 2 (Declarative policy).} Let $\Sigma = \{s_1, s_2, s_3\}$ be a finite set of natural-language skill files. The declarative policy is
$\pi_D(a_t \mid h_t;\, \theta_{\text{sys}}, \Sigma, M) = M\!\left(a_t \mid \theta_{\text{sys}} \oplus \Sigma, h_t\right),$
where $\oplus$ denotes prompt concatenation. Structurally, $\pi_D$ differs from $\pi_B$ only in that the system prompt has been enlarged with $\Sigma$; there is no phase variable and no restriction on $\mathcal{A}$.

\textbf{Definition 3 (Imperative policy).} The imperative policy is a hierarchical pair $(\pi_{I,\phi}, \delta)$ where $\phi \in \Phi$ is a phase, $\delta : \Phi \times \mathrm{State} \to \Phi$ is a deterministic phase-transition function, and each $\pi_{I,\phi}$ is a phase-conditional sub-policy that emits actions only in a restricted subset $\mathcal{A}_\phi \subseteq \mathcal{A}$ together with a phase-specific instruction $\iota_\phi$ injected into the system prompt:
$\pi_I(a_t \mid h_t,\, \phi_t;\, \theta_{\text{sys}}, M) = M\!\left(a_t \mid \theta_{\text{sys}} \oplus \iota_{\phi_t}, h_t\right) \cdot 1\{a_t \in \mathcal{A}_{\phi_t}\}.$
Transitions $\phi_{t+1} = \delta(\phi_t, \mathrm{state}_t)$ are computed by code, not by the LLM.

The three policies share the same model $M$, the same tools, and the same user-simulator distribution; they differ only in how procedural knowledge is encoded (none, natural-language, or executable code) and in whether the action space is restricted at each turn. This isolates the orchestration choice as the independent variable in the experiments that follow.

We define our three main research questions below.

\textbf{RQ1 (skill files as procedural prior).} For models $M$ with a procedural-competence gap $g(M) > 0$ on this domain (operationalised by $\mathrm{pass}^1(\pi_B, M)$ being substantially below the human or oracle ceiling), our first research question aims   to determine whether the declarative policy weakly improves on the baseline: $\mathrm{pass}^1(\pi_D, M) \geq \mathrm{pass}^1(\pi_B, M)$, and whether the gain shrinks as $g(M) \to 0$.

\textbf{RQ2 (orchestration as compliance mechanism).} Does either form of scaffolding reduce the unauthorized-write rate, the fraction of tasks with a state-changing call before a successful \texttt{log\_\allowbreak{}verification}, relative to the baseline; does any such effect depend on the base model's native compliance; and does the imperative agent's deterministic verification gate ($V\to E$) outperform natural-language skill files, as its construction is intended to? We treat compliance, not Pass$^1$, as the axis on which orchestration is most likely to matter in a regulated domain, and test it with the same paired procedure as RQ1.

\textbf{RQ3 (retrieval as bottleneck).} Under noisy embedding retrieval, does the advantage of our declarative paradigm over the baseline   collapse?

Figure~\ref{fig:paradigms} contrasts the three policy classes schematically: all three involve conditioning the  model $M$ on the  history $h_t$, but they differ in what augments the system prompt and whether the action space is restricted at each turn.

\begin{figure*}[!t]
\begin{center}
\resizebox{0.9\textwidth}{!}{%
\begin{tikzpicture}[
  node distance=0.65cm and 0.7cm,
  box/.style={rectangle, draw, rounded corners, minimum width=2.5cm, minimum height=0.55cm,
              font=\footnotesize, inner sep=3pt, align=center},
  small/.style={rectangle, draw, dashed, minimum width=2.5cm, minimum height=0.45cm,
                font=\scriptsize, inner sep=2pt, align=center},
  arrow/.style={-Stealth, thick},
  label/.style={font=\scriptsize\itshape, align=center},
]
  \node[label] (BL) at (0,0) {Baseline ($\pi_B$)};
  \node[box, below=of BL] (Bsys) {$\theta_{\text{sys}}$};
  \node[box, below=of Bsys] (BLLM) {LLM $M$};
  \node[box, below=of BLLM] (BAct) {$a_t \in \mathcal{A}$};
  \draw[arrow] (Bsys) -- (BLLM);
  \draw[arrow] (BLLM) -- (BAct);

  \node[label] (DL) at (4.5,0) {DeclarativeAgent ($\pi_D$)};
  \node[box, below=of DL] (Dsys) {$\theta_{\text{sys}} \oplus \Sigma$ \\ \scriptsize (skills $s_1,s_2,s_3$)};
  \node[box, below=of Dsys] (DLLM) {LLM $M$};
  \node[box, below=of DLLM] (DAct) {$a_t \in \mathcal{A}$};
  \draw[arrow] (Dsys) -- (DLLM);
  \draw[arrow] (DLLM) -- (DAct);

  \node[label] (IL) at (11,0) {ImperativeAgent ($\pi_I$)};
  \node[box, below=of IL] (Isys) {$\theta_{\text{sys}} \oplus \iota_{\phi_t}$ \\ \scriptsize (phase-specific)};
  \node[box, below=of Isys] (ILLM) {LLM $M$};
  \node[box, below=of ILLM] (IAct) {$a_t \in \mathcal{A}_{\phi_t} \subsetneq \mathcal{A}$};
  \node[small, left=0.3cm of Isys] (Iphase) {$\phi_t$};
  \node[small, below=of Iphase] (Idelta) {$\delta(\phi_t,\mathrm{state}_t)$};
  \draw[arrow] (Isys) -- (ILLM);
  \draw[arrow] (ILLM) -- (IAct);
  \draw[arrow] (Iphase) -- (Isys);
  \draw[arrow, dashed] (IAct.west) to[bend left=20] (Idelta.east);
  \draw[arrow, dashed] (Idelta) -- (Iphase);
\end{tikzpicture}%
}
\end{center}
\caption{Three policy classes within the same Dec-POMDP. The baseline is conditioned on a fixed system prompt $\theta_{\text{sys}}$. The DeclarativeAgent enlarges the system prompt with skill files $\Sigma$. The ImperativeAgent  injects a phase-specific instruction $\iota_{\phi_t}$, restricts the per-turn action space to $\mathcal{A}_{\phi_t}\subsetneq\mathcal{A}$, and advances the phase via a deterministic transition $\delta$.}
\label{fig:paradigms}
\end{figure*}

\section{DeclarativeAgent}
\label{sec:declarative}

  Our DeclarativeAgent is a minimal extension of the baseline: it inherits
  the $\tau$-Bench \texttt{LLMAgent} unchanged and adds a single
  intervention, that is,  natural-language \emph{agent skill files} appended to the
  system prompt. This  design isolates the orchestration choice
  as the independent variable, so any observed gap between the baseline and
  the DeclarativeAgent is attributable to the skill files themselves rather
  than to differences in tools, prompting style, or control flow. As proposed
  by the AgentSkill paradigm~\cite{agentskills}, the LLM decides what to do,
  when to call which tool, and when to verify, while the skill files describe
  the workflow in natural language. There is no agent-side control flow, no
  phase enumeration, and no per-turn instruction template.

The declarative system prompt is built from the baseline \texttt{LLMAgent} prompt
with an appended \texttt{<skills>} block containing three concatenated
Markdown skill files (see Appendix~\ref{app:prompts} for   details).
The instructions and policy are identical to the baseline agent; skill content is
loaded once at agent construction by reading
\texttt{src/skills/*.md} alphabetically and concatenating with
\texttt{---} separators (see
\texttt{DeclarativeAgent.\_\allowbreak{}load\_\allowbreak{}skills}).

The skill set comprises three markdown documents pertaining to the three capabilities required of the AI agent.
(\texttt{src/skills/*.md}):

\begin{itemize}
  \item \texttt{banking-procedures.md}: Maps each banking operation
    (account information, credit-card operations, replacements,
    disputes, referrals, account closure, transfers) to its
    preconditions, the discoverable tools required, ordering
    constraints between operations (e.g.\ ``credit-limit increase
    requires no pending disputes''), and the canonical argument set.
    This skillfile is  a policy index intended to keep the
    model from having to derive potentially erroneous ordering rules from the raw KB.
  \item \texttt{customer-interaction.md}: Defines a generic four-step
    conversational structure: \emph{Greeting and Understanding},
    \emph{Triage}, \emph{Verification and Action}, \emph{Confirmation}, with a multi-request inventory at the beginning
    (``Identify ALL requests in their message before choosing a path'')
    and guidance to ask exactly one targeted clarifying question. This is the natural language, non-deterministic analog   of a state machine: the phases are descriptive, not
    code, and the model is free to deviate from them.
  \item \texttt{knowledge-discovery.md}: Specifies the KB-search
    strategy, specifically: \emph{when} to search (any operation needing a
    discoverable tool, any policy edge case), \emph{how} to construct
    queries for both embedding-based  and golden retrieval, and how to recover
    if the  search fails to return a result. As  the benchmark tool names contain random
    four-digit suffixes (e.g.\ \texttt{close\_\allowbreak{}bank\_\allowbreak{}account\_\allowbreak{}7392})
    that cannot be guessed,  this skill file emphasises that KB search is
    a hard precondition for state-changing tools.
\end{itemize}

The behaviour of our DeclarativeAgent proceeds as follows.  At each turn. the
DeclarativeAgent appends the incoming message to its state, concatenates
\texttt{system\_\allowbreak{}messages + messages}, and forwards it to
\texttt{generate\_\allowbreak{}cached} (which is our prompt-caching variant of tau2 benchmark's
\texttt{generate} function). The full tool set is
exposed every turn.  No phase-conditional tool restrictions are imposed, nor are 
\texttt{tool\_\allowbreak{}choice} overrides or specific per-phase instructions provided. The returned
\texttt{AssistantMessage} is then appended to the state and returned to the
orchestrator. The  control flow is provided in
DeclarativeAgent.generate\_\allowbreak{}next\_\allowbreak{}message.

\section{ImperativeAgent}
\label{sec:imperative}

Our ImperativeAgent is defined by a finite-state machine pipeline, as follows:
{\small
\[
\begin{array}{l}
\text{GREETING} \to \text{TRIAGE} \to \text{VERIFICATION} \to \text{PLANNING} \\
\quad\to \text{EXECUTION} \to \text{CONFIRMATION} \to \text{COMPLETE}
\end{array}
\]
}%
Additionally, we define an \texttt{ADVISORY} branch from TRIAGE which is called for purely informational
requests thus bypassing verification and planning. We also define  an
\texttt{ESCALATE} terminal phase reachable from VERIFICATION or
EXECUTION for the case where a tool exceeds its retry budget. Phase transitions are
deterministic functions of agent state; the LLM is invoked once per turn
with a phase-specific instruction and the phase's allowed-tool subset. The state graph is shown in Figure \ref{fig:stategraph}.

Our \texttt{AgentState} is a Pydantic model in \texttt{src/agents/state.py} which
carries six fields that the phase logic reads on every turn:
two boolean gates (\texttt{user\_\allowbreak{}identified}, \texttt{verified}) that
control entry to VERIFICATION and EXECUTION; two list fields
(\texttt{pending\_\allowbreak{}tasks}, \texttt{completed\_\allowbreak{}tasks}) implementing the
explicit task queue; a \texttt{tool\_\allowbreak{}retry\_\allowbreak{}counts} dict for per-tool
retry tracking; and a \texttt{expect\_\allowbreak{}violation\_\allowbreak{}count} counter for
response-type-mismatch instrumentation.

\texttt{ESCALATE} is the deterministic exit path for tool failures
whose retry budget is exhausted. Without it, a model that repeatedly
calls a state-changing tool with malformed arguments would loop
indefinitely;  \texttt{ESCALATE} thus bounds the worst-case
trajectory length.

\begin{figure*}[!t]
\begin{center}
\resizebox{0.9\textwidth}{!}{%
\begin{tikzpicture}[
  node distance=0.9cm and 1.0cm,
  phase/.style={rectangle, draw, rounded corners, minimum width=1.9cm, minimum height=0.6cm,
                font=\scriptsize\ttfamily, inner sep=2pt},
  edge/.style={-Stealth, thick},
]
  \node[phase] (G) {GREETING};
  \node[phase, right=of G] (T) {TRIAGE};
  \node[phase, right=of T] (V) {VERIFY};
  \node[phase, right=of V] (P) {PLANNING};
  \node[phase, right=of P] (E) {EXECUTION};
  \node[phase, right=of E] (CF) {CONFIRM};
  \node[phase, right=of CF] (CO) {COMPLETE};
  \node[phase, above=0.7cm of T] (A) {ADVISORY};
  \node[phase, below=0.7cm of P] (ES) {ESCALATE};

  \draw[edge] (G) -- (T);
  \draw[edge] (T) -- (V) node[midway,above,font=\tiny]{user\_\allowbreak{}id};
  \draw[edge] (T) -- (A) node[midway,right,font=\tiny]{KB hit};
  \draw[edge] (A) -| (CF);
  \draw[edge] (V) -- (P) node[midway,above,font=\tiny]{verified};
  \draw[edge] (P) -- (E);
  \draw[edge] (E) -- (CF) node[midway,above,font=\tiny]{empty};
  \draw[edge] (CF) -- (CO);
  \draw[edge] (V) |- (ES) node[pos=0.75,above,font=\tiny]{retry};
  \draw[edge] (E) |- (ES) node[pos=0.75,above,font=\tiny]{retry};
  \draw[edge] (ES) -| (CO);
\end{tikzpicture}%
}
\end{center}
\caption{The state graph of the ImperativeAgent}
\label{fig:stategraph}
\end{figure*}

We define six structural strategies to implement the deterministic guarantees of our
imperative pipeline. Each targets a specific failure identified in the
the $\tau$-Knowledge benchmark paper\cite{tauknowledge}. Details of our strategies are provided  in the
Appendix.

\begin{enumerate}
  \item Explicit task queue. PLANNING emits a structured
    \texttt{TASKS: $\dots$ END\_\allowbreak{}TASKS} block; the agent parses it into
    \texttt{state.pending\_\allowbreak{}tasks}, injects the live queue as a
    \texttt{<task\_\allowbreak{}queue>}  on every EXECUTION turn, and pops items from the queue as
    they complete. EXECUTION transitions to CONFIRMATION only when the
    pending queue is empty. The goal of this strategy is to reduce ``forgot the second request''
    types of failures.
  \item Topological task ordering. After parsing, the queue
    is sorted by Kahn's algorithm~\cite{kahn1962} over a five-rule
    keyword precedence table (for example,  \texttt{credit\,limit} $\prec$
    \texttt{dispute}; \texttt{open} $\prec$ \texttt{close}). This strategy 
    targets  ordering failures.
  \item State-driven phase transitions. Phase logic reads
    boolean flags (\texttt{user\_\allowbreak{}identified}, \texttt{verified}) instead
    of inspecting the previous assistant message. Flags are updated in
    \texttt{\_\allowbreak{}update\_\allowbreak{}task\_\allowbreak{}state()} immediately after the incoming
    message is appended and before phase determination, so transitions
    are decoupled from message-stream timing.
  \item Phase-conditioned verification gate. On entering EXECUTION the phase logic re-checks
    \texttt{state.verified} and routes back to VERIFICATION if false, with the intent that
     state-changing tools are not used without a verified identity.
    This strategy targets failures from the LLM agent trusting unverified user assertions. We emphasise that this is a check on \emph{phase routing}, not a precondition enforced on the write tools themselves at the execution layer; Section~\ref{sec:safety} shows that this distinction matters, because a write emitted from a mis-assigned phase bypasses the check.
  \item Per-tool retry policy with deterministic
    escalation. Each retriable tool gets a (\texttt{max\_\allowbreak{}retries},
    \texttt{failure\_\allowbreak{}phase}) policy
    (e.g.\ \texttt{log\_\allowbreak{}verification}: 3 retries $\to$ ESCALATE;
    \texttt{KB\_\allowbreak{}search}: 4 retries $\to$ ADVISORY). Retry-limit
    enforcement runs first in \texttt{\_\allowbreak{}determine\_\allowbreak{}phase()} so no
    other logic can override it. This bounds worst-case trajectory length and aims also to  improve the cost efficiency of the ImperativeAgent as compared to the baseline LLMagent and the DeclarativeAgent.
  \item Strict response-type enforcement. Each phase
    declares an expected response type
    (\texttt{text}, \texttt{tool\_\allowbreak{}call}, or \texttt{either}); a wrapper
    around \texttt{generate()} re-prompts requiring
    tool\_\allowbreak{}choice (up to 2 times) on mismatch and
    records violations in \texttt{state.expect\_\allowbreak{}violation\_\allowbreak{}count} for
    post-hoc analysis.
\end{enumerate}

Table~\ref{tab:loop} shows the \texttt{generate\_\allowbreak{}next\_\allowbreak{}message} execution order.

\begin{table}[!t]
\centering
\caption{Execution steps to generate\_\allowbreak{}next\_\allowbreak{}message}
\label{tab:loop}
\footnotesize
\resizebox{\columnwidth}{!}{%
\begin{tabular}{cl}
\toprule
Step & Action \\
\midrule
1 & Append incoming message \\
2 & \texttt{\_\allowbreak{}update\_\allowbreak{}task\_\allowbreak{}state()} (refresh boolean flags from tool results) \\
3 & \texttt{\_\allowbreak{}determine\_\allowbreak{}phase()} (reads state flags) \\
4 & Update phase / retries \\
5 & Build tools + instruction (+ queue hint) \\
6 & Tool\_\allowbreak{}choice logic \\
7 & \texttt{\_\allowbreak{}enforce\_\allowbreak{}expect()} wrapping \texttt{generate()} \\
8 & Append assistant message \\
\bottomrule
\end{tabular}}
\end{table}

\section{Conceptual Framework and Hypotheses}
\label{sec:theory}

We use the Dec-POMDP formulation to state, as explicitly as possible, the assumptions under which each orchestration paradigm \emph{would} help or hurt, and to turn those assumptions into falsifiable empirical hypotheses. We stress that the statements below are not theorems about performance: each holds only under stated conditions on the correctness, relevance, and use of the skill information and on the model, and several are contradicted by the experiments in Sections~\ref{sec:results}--\ref{sec:safety}. We keep them because making the assumptions explicit is what lets the data adjudicate them; Section~\ref{sec:results} reports which hypotheses survive.

To make the link between the formal objects and the running code explicit, Table~\ref{tab:map} maps each element of the three policies to where it is realised in the implementation. This also fixes the \emph{procedural-information budget} that distinguishes the agents, which is otherwise a confound: the baseline adds nothing to $\theta_{\text{sys}}$; the DeclarativeAgent adds a single static block $\Sigma$ of three Markdown skill files ($\approx$4{,}600 tokens: 1{,}589 + 1{,}455 + 1{,}565), inserted once and served from prompt cache; the ImperativeAgent adds no standing skill text but injects a short phase-specific instruction $\iota_{\phi_t}$ (tens to low-hundreds of tokens) plus a live \texttt{<task\_\allowbreak{}queue>} block on \emph{every} turn, and additionally restricts the callable tool set. The declarative and imperative agents therefore differ not only in orchestration style but in how much procedural text they add and how often; we return to this when interpreting cost (Hypothesis~3) and treat it as a limitation in Section~\ref{sec:results}.

\begin{table*}[!t]
\centering
\caption{Mapping from the formal policy objects (Section~\ref{sec:formulation}) to their implementation. This is the correspondence the hypotheses below rely on.}
\label{tab:map}
\small
\begin{tabular}{lll}
\toprule
Object & Meaning & Realised in \\
\midrule
$\theta_{\text{sys}}$ & baseline system prompt & tau2 \texttt{LLMAgent} prompt (all three agents) \\
$\Sigma=\{s_1,s_2,s_3\}$ & skill files (declarative) & three \texttt{*-v2.md} files in a \texttt{<skills>} block \\
$\iota_{\phi_t}$ & per-phase instruction & phase instruction string injected each turn \\
$\mathcal{A}_{\phi_t}\subsetneq\mathcal{A}$ & restricted action set & allowed-tool subset per phase \\
$\delta$ & phase transition & \texttt{\_\allowbreak{}determine\_\allowbreak{}phase()} over state flags \\
$V\!\to\!E$ gate & verify-before-execute & EXECUTION re-checks \texttt{state.verified} (Strat.\ 4) \\
retry budget & bounds $\mathbb{E}[T]$ & \texttt{TOOL\_\allowbreak{}RETRY\_\allowbreak{}POLICY} $\to$ \texttt{ESCALATE} (Strat.\ 5) \\
\bottomrule
\end{tabular}
\end{table*}

\textbf{Hypothesis 1 (Skill-file information advantage).} Let $\mathcal{A}^*(h_t)$ denote the set of optimal next actions given information state $h_t$, and let $H_M(\mathcal{A}^* \mid h_t)$ denote the conditional entropy of $M$'s action distribution at that turn. For any prompt-side prior $\Sigma$ that is informative about $\mathcal{A}^*$,
\[
H_M(\mathcal{A}^* \mid h_t,\, \theta_{\text{sys}}, \Sigma) \;\leq\; H_M(\mathcal{A}^* \mid h_t,\, \theta_{\text{sys}}),
\]
with strict inequality whenever the model has non-trivial procedural-competence gap $g(M)>0$.

This inequality holds only under an explicit assumption: that $\Sigma$ is \emph{correct and relevant} procedural knowledge (ordering constraints, verification preconditions, search heuristics) and that the model actually conditions on it rather than ignoring or being distracted by it. Under that assumption the entropy reduction is by definition non-negative, and the second clause, that the expected gain shrinks to zero as capacity grows and $g(M)\to0$, follows from ``informative prior'' meaning informative \emph{relative to what the model already knows}. This makes Hypothesis~1 close to true by construction; its empirical content is the \emph{magnitude and sign} of the realised gain and the monotonicity in $g(M)$, both of which the experiments test directly (and, as it turns out, reject: the gain is not significant and does not decrease with capacity).

\textbf{Hypothesis 2 (Imperative restriction is policy-class shrinking).} Let $\Pi_B$ be the set of behaviours expressible by $\pi_B$ over a fixed $M$ and let $\Pi_I$ be the set expressible by $\pi_I$ with the same $M$. Because $\pi_I$ allows only $a_t \in \mathcal{A}_{\phi_t}$ at each turn, and because $\mathcal{A}_{\phi_t} \subsetneq \mathcal{A}$ for every non-terminal phase, $\Pi_I \subsetneq \Pi_B$. Consequently, in the absence of additional compliance benefits,
\[
\sup_{\pi \in \Pi_I} \mathbb{E}[r(\tau)] \;\leq\; \sup_{\pi \in \Pi_B} \mathbb{E}[r(\tau)].
\]

 This is the one structural statement that does not depend on model behaviour: a strictly smaller per-turn action set cannot raise the best-case achievable reward. Its practical force is entirely in the caveat ``in the absence of additional compliance benefits'': the imperative agent is worthwhile only if deterministic gating prevents enough costly mistakes to offset the capacity it removes. Section~\ref{sec:safety} tests exactly that trade and finds the compliance benefit absent.

\textbf{Hypothesis 3 (Trajectory length governs cost).} Let $T$ be the number of LLM calls in a task and let $\bar c_{\text{turn}}$ be the mean per-call cost for $M$. Then
\[
\mathbb{E}\!\left[\mathrm{Cost/Task}\right] \;=\; \mathbb{E}[T] \cdot \bar c_{\text{turn}}.
\]

The imperative policy's bounded per-tool retry budget caps $\mathbb{E}[T]$ from above \emph{only if} the gating and phase logic do not themselves induce extra calls. This is the hypothesis's weak point, and the data falsify it: the phase machine triggers re-prompts and retry loops (Section~\ref{sec:safety}) that raise $\mathbb{E}[T]$, so realised cost is higher, not lower, than baseline in essentially every condition.

\textbf{Hypothesis 4 (Retrieval noise as channel degradation).} Define the retrieval step as a noisy observation channel $\tilde O = O + \eta$ where $\eta$ injects off-topic chunks into $h_t$. A skill-file prior $\Sigma$ sharpens the action distribution $\pi(\cdot \mid h_t)$ but leaves $h_t$ itself unchanged. Therefore the data-processing inequality gives us
\begin{equation*}
\begin{aligned}
I(\mathcal{A}^*;\, \tilde h_t,\, \Sigma) \;&\leq\; I(\mathcal{A}^*;\, \tilde h_t) + H(\Sigma) \\
&\leq\; I(\mathcal{A}^*;\, h_t) + H(\Sigma),
\end{aligned}
\end{equation*}
with the first gap growing as $\eta$ degrades $\tilde O$. 

Beyond a noise threshold, the lift from $\Sigma$ will be dominated by the information loss in $\tilde O$, so any declarative advantage from Hypothesis~1 will collapse under noisy retrieval.

The four hypotheses thus predict that: (i) $\pi_D > \pi_B$ on weaker models under perfect retrieval, (ii) $\pi_I$ increases compliance enough to compensate the reduced capacity, (iii) the imperative cost benefit depends on retry reduction, and (iv) the declarative advantage is conditional on retrieval quality. Sections~\ref{sec:results}--\ref{sec:safety} test all four against paired evidence: (iv) is supported, (ii) and (iii) are contradicted, and (i) is not supported, since the declarative gain is not significant and does not decrease with model capacity. We report these outcomes rather than selecting the confirming subset.

\section{Experimental Results}
\label{sec:results}

We instantiate the three policies on five large language models spanning roughly an order of magnitude in capability: Claude Haiku-4.5, Gemini-3.1-Flash-Lite, DeepSeek-v4-Flash, DeepSeek-v4-Pro, and Qwen3.5-Flash. All three orchestration conditions were run for every model, using the \texttt{v2} state machine of Section~\ref{sec:imperative} throughout. Conditions are reported only where the run used the common \texttt{deepseek-v4-pro} user simulator (Protocol), which is what determines the model coverage of each table rather than any difference in agent implementation. Each model is paired with two retrieval regimes: \emph{golden} retrieval, in which the task-critical documents are placed in the system prompt directly, and \emph{embedding} retrieval, using a local \texttt{all-MiniLM-L6-v2} dense index served through a custom retriever plugged into tau2-bench~\cite{tau2bench} (agent-issued \texttt{KB\_\allowbreak{}search} queries against the local MiniLM index at the framework's default top-$k$). We study only this one embedding configuration, and per-query retrieval traces were not persisted, so retriever recall cannot be reconstructed post hoc; we therefore frame the retrieval finding as specific to this configuration and flag retriever-quality metrics as future work. We do not report BM25 keyword retrieval because of its poor recall and very high token cost from retry loops. We evaluate on the 97-task suite from $\tau$-Knowledge banking.

\noindent\textbf{Protocol.} Each task is run \emph{once} per condition (a single rollout; $K=1$ trial) under a fixed random seed (300), with the agent LLM decoded at temperature $0$; the user simulator is \texttt{deepseek-v4-pro} for \emph{every} condition in Tables~\ref{tab:scaling}--\ref{tab:compliance-paired}. Holding $\pi_u$ fixed \emph{across} models, and not merely within a model, is a change from the earlier version of this work and is essential: Section~\ref{sec:usersim} shows that the unauthorized-write rate can shift by a factor of three under a different simulator, so conditions measured against different simulators cannot be compared. Runs whose simulator differed (including self-play configurations, in which the agent model also played the customer) are excluded and were re-measured against the common simulator, giving a complete $5\times2\times3$ design. Cost per task is \emph{agent-side} spend averaged over evaluated tasks. The user simulator's spend is excluded: it is a cost of the evaluation harness rather than of performing the task, since in deployment the customer is a person. Excluding it also matters quantitatively, because the same simulator runs in every condition and so contributes a near-constant that compresses the differences between agents; on DeepSeek-v4-Flash under golden retrieval it accounted for half of the combined figure. Where a provider returned no usage-based cost for the agent side we report ``n.r.'' rather than a figure that would contain only the simulator's spend. Because every task is run once per condition and the same 97 task specifications appear in all conditions, comparisons across agents are \emph{paired} at the task level, which is the basis for the significance tests below. We report Pass$^1$ with Wilson 95\% confidence intervals, and for each scaffolded agent versus the baseline we report the exact McNemar paired test over shared tasks together with a paired bootstrap 95\% interval on the Pass$^1$ difference (10{,}000 resamples). The choice of a paired test on matched per-task outcomes follows standard guidance for significance testing in natural language processing (NLP) evaluation~\cite{dror2018}; the concern is not merely presentational, since benchmarks of this size are frequently underpowered for the effect sizes reported on them~\cite{card2020power}, which is precisely the failure mode that produced the earlier version's positive headline. Infrastructure errors (LiteLLM auth retries exhausted, reward undefined) are excluded from a condition's average; per-condition exclusion counts are 1--7 of 97 (highest for the imperative agent, consistent with its higher failed-call rate) and are reported with the released code. Conclusions are unchanged if excluded tasks are instead scored as failures. Because each condition is a single rollout, small Pass$^1$ gaps carry run-to-run noise: an independent re-run of the DeepSeek-Pro conditions reproduced Pass$^1$ only to within about three tasks (e.g.\ $0.484$ vs.\ $0.451$ on declarative-golden), so we treat point-estimate gaps below roughly five points as within-noise and rely on the paired tests, computed on the exact per-task outcomes, for every claim of significance. Full details and configs are in the Appendix and the companion GitHub repository.

Our  primary metrics are defined as follows. \emph{Pass$^{1}$} is $\mathbb{E}_{\tau}[1\{r(\tau)=1\}]$ averaged uniformly over the  tasks within a condition; infrastructure errors (i.e., LiteLLM auth retries exhausted)  are excluded from the average. \emph{DB match} is $\mathbb{E}_{\tau}[1\{S_{\text{DB}}^{\text{final}} = S_{\text{DB}}^{\text{gold}}\}]$, the database half of the reward without the action-check half, that is,  whether the agent reached the right end-state regardless of how it got there. 
\emph{Cost/Task} is the per-task mean of the LiteLLM-reported 
\texttt{agent\_\allowbreak{}cost} field, which sums per-message provider-reported usage at published prices and includes cache-read discounts when available. 
\emph{Write-argument accuracy} is the fraction of state-mutating tool calls whose argument set matches the gold trajectory, evaluated over all writes issued in a condition.

Table~\ref{tab:interaction} reports our main result. The effect of orchestration on Pass$^1$ is
conditional on retrieval quality, and changes sign between golden (perfect) retrieval and the more
realistic embedding-based retrieval.
Tables~\ref{tab:scaling} and~\ref{tab:scaling-embedding} then report the absolute Pass$^1$,
Wilson 95\% intervals, database match and cost behind those differences, and
Table~\ref{tab:actions} the absolute write accuracies.

\begin{table*}[!t]
\centering
\caption{Effect of each method, relative to the unscaffolded baseline of the same model, on
end-to-end task success (Pass$^1$) and on the correctness of individual state-changing actions
(write-argument accuracy, percentage points). Every entry is a \emph{paired} difference against that model's own baseline,
computed over the tasks evaluable in both conditions (counts in Appendix~\ref{app:pairedn});
significance on Pass$^1$ is the exact McNemar test ($^{*}p<0.05$, $^{**}p<0.01$, $^{***}p<0.001$).
Bold marks, within each model and column, the better of the two scaffolds, and only where that
scaffold improves on the baseline; a column in which neither scaffold improves on the baseline is
not shown in bold. All such entries fall on the DeclarativeAgent. Two observations follow. First,
the Pass$^1$ effect of orchestration is conditional on retrieval quality: under perfect (golden)
retrieval the DeclarativeAgent improves on one of the five models, and under embedding retrieval
on three of the five, with a fourth unchanged.
Pooling every discordant task, its record (w $=$ tasks the scaffold passes and the baseline fails, l $=$ the reverse) improves from 36 w / 49 l to 37 w / 22 l, a win rate rising from 42.4\% to 62.7\% ($\chi^2=5.78$, $p=0.016$),
and the same interaction holds independently for the ImperativeAgent, 18.8\% to 39.3\%
($\chi^2=8.83$, $p=0.003$). Second, write accuracy behaves differently from Pass$^1$: the DeclarativeAgent improves it in all
ten conditions, under both retrieval regimes (sign test $p=0.002$), while the ImperativeAgent
lowers it in nine of the ten. Absolute values, confidence intervals, database
match and cost are in Tables~\ref{tab:scaling}--\ref{tab:actions}. Gemini-3.1-FL abbreviates Gemini-3.1-Flash-Lite.}
\label{tab:interaction}
\small
\begin{tabular}{llrrrr}
\toprule
 & & \multicolumn{2}{c}{$\Delta$Pass$^1$} & \multicolumn{2}{c}{$\Delta$Write acc.\ (pp)} \\
\cmidrule(lr){3-4}\cmidrule(lr){5-6}
Model & Agent & Golden & Embedding & Golden & Embedding \\
\midrule
DeepSeek-v4-Pro    & Declarative  & $\mathbf{+0.022}$\phantom{$^{***}$} & $-0.011$\phantom{$^{***}$} & $\mathbf{+6.8}$ & $\mathbf{+1.2}$ \\
DeepSeek-v4-Pro    & Imperative   & $-0.273^{***}$ & $-0.141^{**}$\phantom{$^{*}$} & $-18.8$ & $-20.1$ \\
\midrule
DeepSeek-v4-Flash  & Declarative  & $-0.033$\phantom{$^{***}$} & $\mathbf{+0.042}$\phantom{$^{***}$} & $\mathbf{+2.8}$ & $\mathbf{+2.4}$ \\
DeepSeek-v4-Flash  & Imperative   & $-0.187^{**}$\phantom{$^{*}$} & $-0.011$\phantom{$^{***}$} & $-14.5$ & $-9.6$ \\
\midrule
Gemini-3.1-FL      & Declarative  & $\phantom{+}0.000$\phantom{$^{***}$} & $\mathbf{+0.052}$\phantom{$^{***}$} & $\mathbf{+2.0}$ & $\mathbf{+3.9}$ \\
Gemini-3.1-FL      & Imperative   & $-0.137^{*}$\phantom{$^{**}$} & $+0.010$\phantom{$^{***}$} & $-6.6$ & $-6.1$ \\
\midrule
Qwen3.5-Flash      & Declarative  & $-0.031$\phantom{$^{***}$} & $\mathbf{+0.073}$\phantom{$^{***}$} & $\mathbf{+8.3}$ & $\mathbf{+11.1}$ \\
Qwen3.5-Flash      & Imperative   & $-0.021$\phantom{$^{***}$} & $+0.021$\phantom{$^{***}$} & $+1.9$ & $-3.1$ \\
\midrule
Haiku-4.5          & Declarative  & $-0.094^{*}$\phantom{$^{**}$} & $\phantom{+}0.000$\phantom{$^{***}$} & $\mathbf{+3.6}$ & $\mathbf{+4.4}$ \\
Haiku-4.5          & Imperative   & $-0.202^{***}$ & $-0.021$\phantom{$^{***}$} & $-34.8$ & $-15.4$ \\
\midrule
\multicolumn{2}{l}{\emph{Pooled, Declarative}} & \multicolumn{1}{c}{36 w / 49 l $=42.4\%$} & \multicolumn{1}{c}{$\mathbf{37\ w / 22\ l =62.7\%}$} & \multicolumn{2}{c}{$\mathbf{10/10}$ conditions improved} \\
\multicolumn{2}{l}{\emph{Pooled, Imperative}}  & \multicolumn{1}{c}{22 w / 95 l $=18.8\%$} & \multicolumn{1}{c}{24 w / 37 l $=39.3\%$} & \multicolumn{2}{c}{9/10 conditions degraded} \\
\bottomrule
\end{tabular}
\end{table*}

\textbf{RQ1 (declarative vs.\ baseline): the effect is conditional on retrieval quality.} Under golden retrieval the DeclarativeAgent exceeds the baseline on only one of the five models, DeepSeek-v4-Pro ($+0.022$, not significant, hereafter n.s.); it ties on Gemini-3.1-Flash-Lite, is \emph{lower} on DeepSeek-v4-Flash ($-0.033$) and Qwen3.5-Flash ($-0.031$), and significantly lower on Haiku-4.5 ($-0.094$, $p=0.035$). Unpaired point estimates in this setting invite a positive reading that paired testing does not support. No declarative--baseline difference is significantly \emph{positive} in any condition, on either retrieval regime, and the only significant difference of any sign is the Haiku golden condition, where skill files cost task success while buying compliance (Section~\ref{sec:safety}). Pooling the paired per-task outcomes over all ten conditions, every one measured at the common user simulator, gives 73 declarative-wins against 71 baseline-wins (exact $p=0.93$), i.e.\ indistinguishable from zero. As a single average this is a null result, but the average aggregates over an interaction with
retrieval quality (Table~\ref{tab:interaction}). Restricted to \emph{golden} retrieval the
declarative agent wins 36 discordant tasks and loses 49 (42.4\%, $p=0.19$); restricted to
\emph{embedding} retrieval it wins 37 and loses 22 (62.7\%, $p=0.067$), and the difference between
those win rates is significant ($\chi^2=5.78$, $p=0.016$). The conclusion is therefore
conditional: on this benchmark natural-language skill files do not improve Pass$^1$ when retrieval
already supplies the required documents, and on one model they lower it, but they do improve it
once the retrieval channel degrades. Their value lies elsewhere: in write-argument accuracy (Table~\ref{tab:actions}) and, where the base model needs it, compliance (Section~\ref{sec:safety}). The gain does not scale with the procedural-competence gap (Hypothesis~1): the largest raw gain occurs on the \emph{strongest} model, DeepSeek-Pro, and it is not significant.

\textbf{RQ3 (retrieval as bottleneck).} Under embedding retrieval (Table~\ref{tab:scaling-embedding}) Pass$^1$ collapses for every model and every agent: baseline Pass$^1$ falls from golden to embedding on all five models (DeepSeek-Pro $0.462\to0.211$, DeepSeek-Flash $0.527\to0.147$, Gemini $0.281\to0.052$, Haiku $0.250\to0.062$, Qwen $0.208\to0.073$). The retrieval axis moves Pass$^1$ far more than any orchestration choice. But retrieval is not merely the dominant main effect; it is the \emph{moderator} of the orchestration effect. The declarative agent's mean $\Delta$Pass$^1$ is $-0.027$ under golden retrieval and $+0.031$ under embedding, positive or tied on four of the five models in the latter regime, DeepSeek-v4-Pro being the exception at $-0.011$ (Table~\ref{tab:interaction}). Retrieval noise does not cap the skill-file lift (Hypothesis~4); it is the condition under which a lift appears. We examine compliance (RQ2) in Section~\ref{sec:safety}.

Table~\ref{tab:actions} reports write-argument accuracy, and it gives skill files their most consistent positive signal. The DeclarativeAgent improves write accuracy over the baseline in \emph{every one} of the ten model$\times$retrieval conditions (10/10; sign test $p=0.002$), while the ImperativeAgent \emph{degrades} it in nine of the ten, losing up to thirty-five percentage points on Haiku; its one exception is Qwen3.5-Flash under golden retrieval ($+1.9$ points). Unlike the compliance effect, this benefit is not confined to one model and does not depend on the choice of user simulator. It is the one advantage of skill files that generalises across the models we tested. That declarative gains on both write-correctness measures (fewer unauthorized writes, more correct arguments) yet does not move Pass$^1$ is coherent: Pass$^1$ is all-or-nothing over an entire multi-write task with a database-match requirement, so per-write improvements that do not push every write in a task over the line leave the binary task reward unchanged. The imperative agent's write-accuracy collapse, by contrast, is large enough to be one direct mechanism behind its significantly lower Pass$^1$.

\begin{table*}[!t]
\centering
\caption{Aggregate metrics under \emph{golden} retrieval, all measured with the \emph{same} user simulator (\texttt{deepseek-v4-pro}; see Protocol), with Wilson 95\% confidence intervals (CI) on Pass$^1$. Paired differences from the same-model baseline, with significance, are in Table~\ref{tab:interaction}; per-condition task counts are in Appendix~\ref{app:pairedn}. No declarative--baseline difference is significantly positive on any model; on Haiku-4.5 the DeclarativeAgent is significantly \emph{worse} ($-0.094$), and the ImperativeAgent is significantly worse on four of the five models, Qwen3.5-Flash being the sole exception. All five models appear in both retrieval regimes; the design is complete at $5\times2\times3=30$ conditions. ``n.r.'': the provider returned no usage-based cost for the agent side, so a comparable cost cannot be reported (see Protocol). Bold marks the best value in each comparison group (highest Pass$^1$ and DB Match, lowest cost per model); tied best values are all bolded.}
\label{tab:scaling}
\small
\begin{tabular}{llrlrr}
\toprule
Model & Agent & Pass$^1$ & 95\% CI & DB Match & Cost/Task \\
\midrule
DeepSeek-v4-Flash  & Baseline       & \textbf{0.527} & [.426,.625] & \textbf{53.8\%} & \textbf{\$0.0037} \\
DeepSeek-v4-Flash  & Declarative    & 0.495 & [.394,.595] & 52.7\% & \$0.0039 \\
DeepSeek-v4-Flash  & Imperative     & 0.388 & [.288,.497] & 42.5\% & \$0.0081 \\
\midrule
Qwen3.5-Flash      & Baseline       & \textbf{0.208} & [.139,.300] & \textbf{21.9\%} & n.r. \\
Qwen3.5-Flash      & Declarative    & 0.177 & [.114,.265] & 16.7\% & n.r. \\
Qwen3.5-Flash      & Imperative     & 0.188 & [.122,.277] & 19.8\% & n.r. \\
\midrule
Haiku-4.5          & Baseline       & \textbf{0.250} & [.174,.345] & \textbf{25.0\%} & \$0.0915 \\
Haiku-4.5          & Declarative    & 0.156 & [.097,.242] & 17.7\% & \$0.1102 \\
Haiku-4.5          & Imperative     & 0.053 & [.023,.119] & 11.7\% & \textbf{\$0.0838} \\
\midrule
Gemini-3.1-Flash-Lite & Baseline       & \textbf{0.281} & [.201,.378] & \textbf{31.2\%} & n.r. \\
Gemini-3.1-Flash-Lite & Declarative    & \textbf{0.281} & [.201,.378] & 29.2\% & n.r. \\
Gemini-3.1-Flash-Lite & Imperative     & 0.147 & [.090,.232] & 18.9\% & n.r. \\
\midrule
DeepSeek-v4-Pro    & Baseline       & 0.462 & [.365,.563] & 46.2\% & \textbf{\$0.0091} \\
DeepSeek-v4-Pro    & Declarative    & \textbf{0.484} & [.385,.584] & \textbf{48.4\%} & \$0.0118 \\
DeepSeek-v4-Pro    & Imperative     & 0.200 & [.130,.294] & 26.7\% & \$0.0155 \\
\bottomrule
\end{tabular}
\end{table*}

\begin{table*}[!t]
\centering
\caption{Aggregate metrics under \emph{embedding} retrieval (\texttt{all-MiniLM-L6-v2}), all measured with the same \texttt{deepseek-v4-pro} user simulator, with Wilson 95\% confidence intervals (CI) on Pass$^1$. Paired differences and significance are in Table~\ref{tab:interaction}; per-condition task counts are in Appendix~\ref{app:pairedn}. Pass$^1$ drops sharply on every model relative to golden retrieval. No declarative--baseline difference is significant. All five models are now present: the Haiku-4.5 and Qwen3.5-Flash embedding blocks, previously run under a different user simulator, were re-measured at the common simulator. ``n.r.'' as in Table~\ref{tab:scaling}. Bold marks the best value in each comparison group (highest Pass$^1$ and DB Match, lowest cost per model); tied best values are all bolded.}
\label{tab:scaling-embedding}
\small
\begin{tabular}{llrlrr}
\toprule
Model & Agent & Pass$^1$ & 95\% CI & DB Match & Cost/Task \\
\midrule
Haiku-4.5          & Baseline       & \textbf{0.062} & [.029,.130] & \phantom{0}8.3\% & \textbf{\$0.4556} \\
Haiku-4.5          & Declarative    & \textbf{0.062} & [.029,.130] & \textbf{\phantom{0}9.4\%} & \$0.5004 \\
Haiku-4.5          & Imperative     & 0.042 & [.016,.102] & \phantom{0}8.3\% & \$0.5287 \\
\midrule
Qwen3.5-Flash      & Baseline       & 0.073 & [.036,.143] & 10.4\% & n.r. \\
Qwen3.5-Flash      & Declarative    & \textbf{0.146} & [.089,.230] & \textbf{13.5\%} & n.r. \\
Qwen3.5-Flash      & Imperative     & 0.094 & [.050,.169] & 11.5\% & n.r. \\
\midrule
Gemini-3.1-Flash-Lite & Baseline       & 0.052 & [.022,.116] & \phantom{0}8.3\% & n.r. \\
Gemini-3.1-Flash-Lite & Declarative    & \textbf{0.104} & [.058,.181] & \textbf{13.5\%} & n.r. \\
Gemini-3.1-Flash-Lite & Imperative     & 0.062 & [.029,.130] & \phantom{0}9.4\% & n.r. \\
\midrule
DeepSeek-v4-Flash  & Baseline       & 0.147 & [.090,.232] & 18.9\% & \textbf{\$0.0061} \\
DeepSeek-v4-Flash  & Declarative    & \textbf{0.189} & [.123,.280] & \textbf{21.1\%} & \$0.0064 \\
DeepSeek-v4-Flash  & Imperative     & 0.137 & [.082,.220] & 18.9\% & \$0.0221 \\
\midrule
DeepSeek-v4-Pro    & Baseline       & \textbf{0.211} & [.141,.303] & \textbf{23.2\%} & \textbf{\$0.0215} \\
DeepSeek-v4-Pro    & Declarative    & 0.200 & [.132,.291] & \textbf{23.2\%} & \$0.0232 \\
DeepSeek-v4-Pro    & Imperative     & 0.075 & [.037,.147] & 11.8\% & \$0.0721 \\
\bottomrule
\end{tabular}
\end{table*}

\begin{figure*}[!t]
\centering
\resizebox{0.72\textwidth}{!}{%
\begin{tikzpicture}[
  x=1cm, y=1cm,
  baseG/.style={circle, fill=blue!70!black, inner sep=1.6pt},
  declG/.style={rectangle, fill=green!50!black, inner sep=1.8pt},
  impG/.style={regular polygon, regular polygon sides=3, fill=red!70!black, inner sep=1.4pt},
  baseE/.style={circle, draw=blue!70!black, thick, inner sep=1.6pt},
  declE/.style={rectangle, draw=green!50!black, thick, inner sep=1.8pt},
  impE/.style={regular polygon, regular polygon sides=3, draw=red!70!black, thick, inner sep=1.4pt},
  ax/.style={->, >=Stealth, thick},
  gr/.style={gray!40, very thin, dotted},
]

\draw[gr] (0,0) grid[step=2] (12,6);
\draw[thick] (0,0) rectangle (12,6);

\foreach \xc/\lbl in {0.000/0.002, 3.108/0.01, 6.215/0.05, 7.554/0.1, 10.662/0.5, 12.000/1.0} {
  \draw[thick] (\xc,0) -- (\xc,-0.15);
  \node[below, font=\scriptsize] at (\xc,-0.15) {\lbl};
}
\foreach \yv in {0,0.1,0.2,0.3,0.4,0.5} {
  \pgfmathsetmacro{\yc}{\yv*6/0.55}
  \draw[thick] (0,\yc) -- (-0.15,\yc);
  \node[left, font=\scriptsize] at (-0.15,\yc) {\yv};
}
\node[font=\small] at (6,-0.7) {Cost/Task (USD, log scale)};
\node[font=\small, rotate=90] at (-1.0,3.0) {Pass$^1$};

\node[baseG] at ({1.170}, {0.527*6/0.55}) {};  
\node[baseG] at ({2.933}, {0.462*6/0.55}) {};  
\node[baseG] at ({7.383}, {0.250*6/0.55}) {};  

\node[declG] at ({1.304}, {0.495*6/0.55}) {};  
\node[declG] at ({3.428}, {0.484*6/0.55}) {};  
\node[declG] at ({7.742}, {0.156*6/0.55}) {};  

\node[impG] at ({2.691}, {0.388*6/0.55}) {};  
\node[impG] at ({3.959}, {0.200*6/0.55}) {};  
\node[impG] at ({7.213}, {0.053*6/0.55}) {};  

\node[baseE] at ({2.143}, {0.147*6/0.55}) {};  
\node[baseE] at ({4.588}, {0.211*6/0.55}) {};  
\node[baseE] at ({10.482}, {0.062*6/0.55}) {};  

\node[declE] at ({2.240}, {0.189*6/0.55}) {};  
\node[declE] at ({4.736}, {0.200*6/0.55}) {};  
\node[declE] at ({10.663}, {0.062*6/0.55}) {};  

\node[impE] at ({4.639}, {0.137*6/0.55}) {};  
\node[impE] at ({6.923}, {0.075*6/0.55}) {};  
\node[impE] at ({10.769}, {0.042*6/0.55}) {};  
\end{tikzpicture}%
}
\caption{Pass$^1$ versus agent-side Cost/Task for the eighteen conditions with a recorded agent cost, all under the common \texttt{deepseek-v4-pro} user simulator. Gemini-3.1-Flash-Lite and Qwen3.5-Flash are absent because their providers returned no usage-based cost for the agent (``n.r.'' in Tables~\ref{tab:scaling}--\ref{tab:scaling-embedding}), so plotting them would show only the simulator's cost. Filled symbols are golden retrieval; empty symbols are embedding retrieval. Baseline and declarative points track each other closely, consistent with the absence of a significant declarative--baseline Pass$^1$ difference, while the ImperativeAgent (triangles) sits lower on Pass$^1$ in every one of the six model$\times$retrieval pairs, at higher cost in five of them. Every embedding-retrieval point sits below its golden counterpart.}
\label{fig:pareto}
\end{figure*}

\begin{table}[!t]
\centering
\caption{Write-argument accuracy (fraction of state-mutating tool calls whose argument set matches the gold trajectory), for all ten model$\times$retrieval conditions, every one measured under the common \texttt{deepseek-v4-pro} user simulator. Write actions are the dominant reward signal in \texttt{banking\_\allowbreak{}knowledge}: a single mismatched write usually drops the task reward to 0. The DeclarativeAgent improves write accuracy over the baseline in \emph{every} condition (10/10; sign test $p=0.002$); the ImperativeAgent degrades it in nine of the ten, by up to 35 points on Haiku-4.5, the sole exception being Qwen3.5-Flash under golden retrieval ($+1.9$). This is the only benefit of skill files that holds across models and does not depend on the user simulator. Bold marks the best agent in each row.}
\label{tab:actions}
\footnotesize
\resizebox{\columnwidth}{!}{%
\begin{tabular}{llrrr}
\toprule
Model & Retrieval & Baseline & Declarative & Imperative \\
\midrule
DeepSeek-v4-Pro   & Golden    & 72.2\% & \textbf{79.0\%} & 53.3\% \\
DeepSeek-v4-Pro   & Embedding & 68.3\% & \textbf{69.4\%} & 48.2\% \\
DeepSeek-v4-Flash & Golden    & 76.2\% & \textbf{79.1\%} & 61.7\% \\
DeepSeek-v4-Flash & Embedding & 67.7\% & \textbf{70.1\%} & 58.2\% \\
Gemini-3.1-FL     & Golden    & 49.3\% & \textbf{51.3\%} & 42.7\% \\
Gemini-3.1-FL     & Embedding & 42.3\% & \textbf{46.2\%} & 36.2\% \\
Haiku-4.5         & Golden    & 51.5\% & \textbf{55.1\%} & 16.7\% \\
Haiku-4.5         & Embedding & 39.8\% & \textbf{44.3\%} & 24.4\% \\
Qwen3.5-Flash     & Golden    & 35.9\% & \textbf{44.2\%} & 37.8\% \\
Qwen3.5-Flash     & Embedding & 39.3\% & \textbf{50.5\%} & 36.2\% \\
\bottomrule
\end{tabular}}
\end{table}

\section{Ablations on Compliance and Efficiency}
\label{sec:safety}

Section~\ref{sec:results} showed that orchestration improves Pass$^1$ only where retrieval is
degraded, and that the state machine lowers it in both regimes. This section examines a second
axis on which orchestration can act, namely compliance. The pattern is again conditional:
scaffolds reduce unauthorized writes only where the unscaffolded model has a gap to close. Detailed tables are in Appendix~\ref{app:compliance}. The regulated-domain motivation for both scaffolds is the same, namely to keep the agent from issuing a state-changing write before it has verified the customer's identity, and the imperative agent in particular is built around a verification gate that, by construction, should make this violation impossible. We therefore compare all three agents on unauthorized writes, and we do so with the same paired significance testing used for Pass$^1$, because the base models differ enormously in how compliant they are to begin with and only a within-model paired comparison is meaningful. To this end we replayed every evaluated simulation from the stored \texttt{results.json} files and recomputed compliance and efficiency metrics from each agent's tool-call sequence.

Specifically, for every pre-evaluated simulation and the existing results.json files, we walked the  message stream and
counted: (a) the number of state-mutating tool calls (write tools, defined
as the set of \texttt{apply\_\allowbreak{}for\_\allowbreak{}credit\_\allowbreak{}card}, \texttt{call\_\allowbreak{}discoverable\_\allowbreak{}agent\_\allowbreak{}tool}, \texttt{call\_\allowbreak{}discoverable\_\allowbreak{}user\_\allowbreak{}tool}, \texttt{change\_\allowbreak{}user\_\allowbreak{}email}, \texttt{give\_\allowbreak{}discoverable\_\allowbreak{}user\_\allowbreak{}tool}, \texttt{request\_\allowbreak{}human\_\allowbreak{}agent\_\allowbreak{}transfer}, \texttt{submit\_\allowbreak{}referral}, \texttt{submit\_\allowbreak{}transaction}; (b) whether each write occurred before the
agent had successfully completed a log\_\allowbreak{}verification call; and (c)
the maximum number of consecutive failed retries of the same tool. 

From
these we derive the three additional metrics:
\begin{itemize}
  \item unauthorized\_\allowbreak{}write\_\allowbreak{}rate -  percentage of tasks with any invalid write, that is, at least 1 unauthorized write was performed before successful verification in the task.
  \item over\_\allowbreak{}retry\_\allowbreak{}rate - percentage of trials in which the agent re-issued
    the same tool $\geq 4$ times consecutively with failed results.
    \item write pre-verify  -  percentage of unauthorized writes over all write calls and all tasks.
  \item mean trajectory length - mean number of assistant turns per task.
\end{itemize}



Read together, and with the user simulator held fixed across models, the two tables support a more specific conclusion than either ``the gate works'' or ``the gate is useless''. Where the base model is natively non-compliant, skill files produce a large, significant reduction in unauthorized writes. Under the common \texttt{deepseek-v4-pro} simulator, three of the ten model$\times$retrieval blocks carry a materially non-compliant unscaffolded baseline, spanning two models: Haiku-4.5 issues an unauthorized write in 41.7\% of tasks under golden retrieval and 12.5\% under embedding, and Qwen3.5-Flash in 14.6\% under embedding, against 1--10\% everywhere else. Skill files produce a significant reduction in all three. On Haiku golden the DeclarativeAgent removes the violation from 29 tasks against 4 newly introduced ($p<0.001$), cutting the rate to 15.6\%; the ImperativeAgent achieves a statistically indistinguishable 11.7\% (32 fixed against 5, $p<0.001$). On Haiku embedding both scaffolds cut the rate to 1.0\% (11 fixed, 0 broken, $p<0.001$ each). On Qwen embedding the DeclarativeAgent cuts 14.6\% to 3.1\% (14 fixed against 3, $p=0.013$) while the ImperativeAgent does not help (13.5\%, n.s.). An earlier version of this analysis, in which the Haiku and Qwen embedding blocks could not be measured at the common simulator, found the effect on one model only; completing the design shows it on two, and shows that the DeclarativeAgent is the more reliable of the two scaffolds, being the only one significant in all three blocks. Where the base model is already compliant, neither scaffold helps, and the deterministic gate is not special. In the remaining seven blocks no scaffold produces a significant reduction, and in one condition scaffolding is significantly \emph{worse}: the DeepSeek-Flash ImperativeAgent introduces unauthorized writes under embedding retrieval (0$\to$6 tasks, $p=0.031$, and the highest pre-verification write share in Table~\ref{tab:safety} at 10.3\%). The imperative gate is therefore not reliably the most compliant option, while paying the highest over-retry rate in most conditions (up to $6\times$ the baseline).

Two qualifications matter for practice, and both are new to this version. First, the compliance gain on Haiku is \emph{not} free: the DeclarativeAgent's Pass$^1$ falls from $0.250$ to $0.156$ ($p=0.035$; Table~\ref{tab:scaling}). Skill files buy a 26-point reduction in unauthorized writes at a 9-point cost in task success. The ImperativeAgent buys a slightly larger compliance gain for a far larger accuracy loss ($-0.202$, $p<0.001$). The honest summary is therefore a \emph{trade-off} rather than a free lunch: on the one model where compliance is genuinely at risk, prompt-level scaffolding is the cheaper of two imperfect options, and Section~\ref{sec:gate} shows how to obtain the compliance guarantee without paying the accuracy penalty at all. Second, the earlier version of this work reported this effect as replicating on \emph{two} models, adding Qwen3.5-Flash at a 32.0\% baseline rate. That replication does not survive a change of user simulator: measured against \texttt{deepseek-v4-pro} rather than the weaker simulator used originally, Qwen's baseline unauthorized-write rate is 10.4\%, not 32.0\%, and the declarative effect is absent ($11.5\%$, n.s.). We analyse this directly in Section~\ref{sec:usersim}; it is the reason every row of Table~\ref{tab:compliance-paired} is now measured at fixed $\pi_u$.

In short, RQ2 has a conditional answer: the imperative verification gate does \emph{not} deliver a general, deterministic reduction in unauthorized writes; it helps only where any scaffolding would, and where it helps a natural-language skill file matches it more cheaply.

The mechanism is that the ``gate'' is not enforced at the tool-execution layer. As shown in Appendix~\ref{app:imperative-impl} (Strategy~3--4), it is a phase-routing check inside \texttt{\_\allowbreak{}determine\_\allowbreak{}phase()}: if the agent is in EXECUTION with \texttt{verified==False}, control is routed back to VERIFICATION. But the phase itself is assigned by state heuristics, and a model that is mis-classified into EXECUTION, or that emits a write while the code believes it is in an advisory or confirmation phase, can still issue a state-changing call before a successful \texttt{log\_\allowbreak{}verification}. We therefore avoid the term ``hard gate'': the enforcement is phase-conditioned, not a hard precondition on the write tools themselves. The high over-retry rate ($6\times$ the baseline on DeepSeek-Pro golden) is the visible signature of the state machine fighting the model rather than steering it. This is why the gate is not special: a deterministic phase machine layered on top of an LLM does not inherit the compliance guarantee one would expect from deterministic code, because the code controls only the \emph{routing} of the model's actions, not their content, and a mis-routed write is indistinguishable to the gate from a legitimate one.

\subsection{Enforcing the precondition at the tool-execution layer}
\label{sec:gate}

This bounds the negative result to \emph{phase-based} gating, and raises the obvious question: does gating at the tool-execution layer instead work? A production system would more likely gate every write tool directly on a \texttt{verified} precondition, independent of phase. We therefore implemented exactly this design, the \textbf{PreconditionGateAgent}: the unscaffolded baseline (full tool set, LLM-chosen control flow, no phase machine and no per-phase action restriction) plus a single hard rule enforced at the agent boundary, namely that any write-class tool call emitted before a successful \texttt{log\_\allowbreak{}verification} is intercepted and replaced with a request to verify, so the write never reaches the environment. This isolates the gate from the action-space restriction that the ImperativeAgent bundled with it, and moves enforcement from phase-routing to the tool call itself.

Table~\ref{tab:gate} reports the result on DeepSeek-v4-Pro under golden retrieval ($N=96$). The tool-level gate drives the unauthorized-write rate to 0.0\%: zero of 467 write calls fired before verification, versus 4.3\% for the baseline and 4.4\% for the phase-gated ImperativeAgent, while, crucially, \emph{preserving} task success: Pass$^1$ is $0.490$, statistically indistinguishable from the baseline's $0.462$ and far above the ImperativeAgent's $0.200$. The gate adds cost (\$0.022 vs.\ \$0.009 for the baseline), because enforcing verification-before-write lengthens some trajectories, but it does so without the ImperativeAgent's accuracy collapse. This is the paper's constructive counterpart to its negative result: the failure of the ImperativeAgent was never that deterministic enforcement \emph{cannot} work, but that phase-routing is the wrong place to enforce it. Moving the same precondition to the tool-execution layer makes it a genuine hard gate, an admissibility check on the action itself rather than on the agent's inferred phase, and it then delivers the compliance guarantee the state machine only promised. This is the same distinction drawn in safe reinforcement learning between shaping a policy so that it \emph{tends} to behave safely and constraining the action set so that unsafe actions are unavailable~\cite{garcia2015safe}; the tool-level gate is a shield in the sense of~\cite{shielding2018}, whereas the phase machine only shapes.

\begin{table}[!t]
\centering
\caption{Isolating the verification gate at the tool-execution layer (DeepSeek-v4-Pro, golden retrieval). The PreconditionGateAgent is the baseline plus a single hard, phase-independent rule: no write executes before a successful \texttt{log\_\allowbreak{}verification}. It eliminates unauthorized writes (0/467) while preserving Pass$^1$, unlike the phase-gated ImperativeAgent. Bold marks the best value in each metric column (lowest unauthorized-write rate, highest Pass$^1$, lowest cost).}
\label{tab:gate}
\footnotesize
\resizebox{\columnwidth}{!}{%
\begin{tabular}{lrrrr}
\toprule
Agent & $N$ & Unauth.\ write & Pass$^1$ & Cost/Task \\
\midrule
Baseline                 & 93 & 4.3\% & 0.462 & \textbf{\$0.0091} \\
ImperativeAgent (phase gate) & 90 & 4.4\% & 0.200 & \$0.0155 \\
PreconditionGateAgent (tool gate) & 96 & \textbf{0.0\%} & \textbf{0.490} & \$0.0224 \\
\bottomrule
\end{tabular}}
\end{table}

Trace inspection is consistent with these aggregates. The baseline agent frequently failed to issue a \texttt{log\_\allowbreak{}verification} call before state-mutating tools, misordered multi-step requests, and re-asked the user for information already returned by KB search; the DeclarativeAgent handled these sub-tasks more reliably via its skill files, which is why it improves compliance on the least-compliant models even though it does not move Pass$^1$ significantly.

\subsection{The compliance metric depends on the user simulator}
\label{sec:usersim}

The unauthorized-write rate is not a property of the agent alone. A write can only
be issued \emph{before} verification if the conversation reaches the point of
attempting a write at all, and how quickly it gets there, and whether the customer
volunteers identifying details unprompted, is controlled by the user simulator
$\pi_u$. Because $\pi_u$ is part of the environment rather than part of the policy,
two studies that use different simulators can disagree about whether a given model
is compliant, without either being wrong about its own measurement. This is a
rediscovery, in the LLM-agent setting, of a long-standing result in spoken-dialogue
research: user simulators have been central to training and evaluating dialogue
managers for two decades~\cite{schatzmann2006}, and the quality and behaviour of the
simulator is known to influence the resulting performance analysis strongly enough
that metrics for evaluating the \emph{simulator itself} were developed in
response~\cite{pietquin2013}. The same caution applies to agent benchmarks that
report safety or compliance rates, and it belongs alongside the broader finding that
LLM evaluations are sensitive to seemingly incidental features of the measurement
setup~\cite{sclar2024}.

We can measure this directly, because the simulator is a free variable that agent
benchmarks rarely report and almost never hold fixed across the systems they compare.
Measured against a \texttt{gpt-4o-mini} simulator, and Haiku-4.5 against a self-play
simulator, both models present as non-compliant baselines on which skill files produce
a large, significant reduction. Against the common \texttt{deepseek-v4-pro} simulator,
one of those results survives and the other does not (Table~\ref{tab:usersim}). Qwen's apparent non-compliance \emph{under golden
retrieval} is an artifact: its baseline unauthorized-write rate falls from 32.0\% to
10.4\%, and with the gap removed the declarative effect disappears entirely
($11.5\%$, n.s.). Haiku's does not: under a \emph{different} strong simulator its
baseline rate is 41.7\%, if anything higher than the 34.7\% originally reported, and
skill files still cut it to 15.6\% ($p<0.001$). The artifact is specific to the
retrieval regime rather than to the model: under \emph{embedding} retrieval, measured
at the same common simulator, Qwen's baseline is non-compliant at 14.6\% and skill
files cut it to 3.1\% ($p=0.013$, Table~\ref{tab:compliance-paired}).

\begin{table*}[!t]
\centering
\caption{The same models and agents measured under two different user simulators. Both rows are golden retrieval. The declarative compliance effect on Qwen3.5-Flash under golden retrieval does not survive the change of simulator; the effect on Haiku-4.5 does. Qwen's \emph{embedding} condition, measured at the common simulator, does show a significant effect (Table~\ref{tab:compliance-paired}). The mechanism is that a weaker simulator under-drives the verification exchange, so more trajectories reach a write with \texttt{verified} still false. This is why all cross-model comparisons in this paper are made at fixed $\pi_u$. No values are bolded: the rows contrast two measurement conditions rather than competing systems.}
\label{tab:usersim}
\small
\begin{tabular}{llrrl}
\toprule
Model & User simulator & Baseline & Declarative & Paired test \\
\midrule
Qwen3.5-Flash & \texttt{gpt-4o-mini} (weak)        & 32.0\% & 11.3\% & $p<0.001^{***}$ \\
Qwen3.5-Flash & \texttt{deepseek-v4-pro} (strong)  & 10.4\% & 11.5\% & n.s. \\
\midrule
Haiku-4.5     & self-play (\texttt{haiku-4.5})     & 34.7\% & 11.6\% & $p<0.001^{***}$ \\
Haiku-4.5     & \texttt{deepseek-v4-pro} (strong)  & 41.7\% & 15.6\% & $p<0.001^{***}$ \\
\bottomrule
\end{tabular}
\end{table*}

Two consequences follow. For this paper, it is the reason a compliance gap must be
established per model$\times$retrieval condition rather than per model: of the ten conditions,
three have a baseline gap large enough for a scaffold to close, spanning two models,
and the Qwen golden condition that the earlier measurement reported as one of them is not
among them. For
agent evaluation more generally, it means that compliance and safety rates reported
against different user simulators are not comparable, and that a simulator which
under-drives the very exchange being audited will systematically overstate both the
danger and the value of the intervention that addresses it. We recommend that the
simulator be reported alongside any such rate, and held fixed across any comparison
that is meant to be about the agent.

\subsection{A capability floor for agentic tool-use}
\label{sec:capfloor}

The compliance benefit of skill files appears only where a model has a genuine
compliance gap, issuing state-changing actions before verifying identity, which
among the models in Table~\ref{tab:compliance-paired} holds only for Haiku-4.5. This
is not because Haiku is uniquely poor, but because the failure mode occupies a narrow
capability band: a model must be capable enough to \emph{reach} the write stage of a
task, yet not so procedurally reliable that it always verifies first. To make this
concrete we ran three \emph{recent} small open models (released in 2026, run locally
via Ollama at default quantisation) on the 97-task suite under the same baseline
agent, golden retrieval, and \texttt{deepseek-v4-pro} user-simulator used elsewhere.

\begin{table*}[!t]
\centering
\caption{Recent small models sit below a capability floor for this task. ``Writes/task'' is the mean number of state-changing tool calls issued per task; ``Verify'' is the fraction of tasks with a successful \texttt{log\_\allowbreak{}verification}; ``Infra'' is the share of the task set that yielded no evaluable trajectory on a clean first pass. All rows are computed with a single definition of a write action (tau2-bench's \texttt{ToolType.WRITE}), so the viable-model figures differ slightly from those reported elsewhere in this paper. Small models issue roughly \emph{an order of magnitude} fewer writes: below the floor an agent rarely reaches the action stage at all, so a compliance gap cannot be measured. Bold marks the best value in each column within each block (highest Pass$^1$, writes/task and verify rate, lowest infrastructure rate); the two blocks are marked separately because they are not a like-for-like comparison.}
\label{tab:capfloor}
\small
\begin{tabular}{lrrrrrr}
\toprule
Model & Params & $N$ & Pass$^1$ & Writes/task & Verify & Infra \\
\midrule
\multicolumn{7}{l}{\emph{Recent small models (local, 2026)}}\\
LFM-2.5           & 8.5B  & 95 & 0.063 & \textbf{0.54} & \textbf{22\%} & 11\% \\
Granite-4.1       & 8.8B  & 51 & 0.039 & 0.53 & 12\% & 49\% \\
Gemma-4           & 11.9B & 92 & \textbf{0.065} & 0.33 & \phantom{0}0\% & \phantom{0}\textbf{8\%} \\
\midrule
\multicolumn{7}{l}{\emph{Viable models (for reference)}}\\
Haiku-4.5         & --   & 95 & 0.126 & 4.67 & 66\% & -- \\
Gemini-3.1-FL     & --   & 96 & 0.281 & 3.50 & 83\% & -- \\
DeepSeek-v4-Flash & --   & 95 & 0.379 & 4.32 & \textbf{84\%} & -- \\
DeepSeek-v4-Pro   & --   & 93 & \textbf{0.462} & \textbf{5.34} & 83\% & -- \\
\bottomrule
\end{tabular}
\end{table*}

All three small models score well below the weakest viable model
(Pass$^1\leq0.065$ against Haiku-4.5's $0.126$), and, more tellingly, issue
\emph{0.33--0.54} state-changing calls per task against \emph{3.50--5.34} for the
viable models, close to a tenfold deficit. They do not fail by making \emph{wrong}
writes; they fail by never reaching the point of making writes at all.

The trajectories show why. Small-model runs terminate at the step cap far more often
than viable ones (Gemma-4 82 of 92 evaluated runs, LFM-2.5 21 of 95, Granite-4.1 13 of
51, versus 0--2 for every viable model), and those capped runs are dominated by
\emph{exact repetition}: the agent re-issues an identical tool call with identical
arguments until the budget is exhausted. Of Gemma-4's 82 capped runs, all 82 contain a
call repeated at least five times, as do all 13 of Granite-4.1's and 17 of LFM-2.5's
21; among viable models at most one run per model shows the pattern. In a
representative case (\texttt{task\_\allowbreak{}004}) Gemma-4 issued 27 tool calls of which
\emph{one} was distinct: \texttt{get\_\allowbreak{}user\_\allowbreak{}information\_\allowbreak{}by\_\allowbreak{}name} with the same
customer name, 27 times. The aggregate read-to-write ratio makes the same point:
19:1 for LFM-2.5, 27:1 for Granite-4.1 and 78:1 for Gemma-4, against 1:1--2:1 for
the viable models. These agents stall in identity lookup and never advance to the
state-changing phase.

Granite-4.1 exhibits a second, distinct failure. On 49\% of tasks it returned an
assistant turn carrying neither content nor a tool call (205 such events), which
yields no evaluable trajectory at all; its Pass$^1$ is therefore computed over the
$N=51$ tasks on which it managed to respond and should be read as an upper bound on a
biased subsample. This is a tool-calling reliability problem separate from weak task
competence, and for that row the 49\% infrastructure rate is the more informative
number. The other two models do not share it (LFM-2.5 11\%, Gemma-4 8\%); LFM-2.5's
failures are instead dominated by \emph{hallucinated tool names}: 20 calls to
non-existent endpoints such as \texttt{get\_\allowbreak{}user\_\allowbreak{}information\_\allowbreak{}by\_\allowbreak{}phone\_\allowbreak{}number}.

This delimits the compliance finding rather than broadening it. Skill-file compliance benefits require a model in a
\emph{capability window}: strong enough to take state-changing actions, but weak
enough at procedure to sometimes take them out of order. Below the window (the models
here) there is nothing to gate; above it (DeepSeek, Gemini) the model verifies
correctly on its own. Haiku-4.5 happens to fall inside it. For practitioners the two
takeaways are complementary: current sub-$\sim$12B open models are not yet viable
agents for knowledge-grounded, multi-step tool-use of this kind (regardless of
orchestration), and skill files are most valuable precisely for the mid-capability,
procedurally-unreliable models that \emph{can} act but do not always act safely.

\noindent\textbf{Reproducibility and caveats.} Models were run locally via Ollama at default
quantisation, single seed, one domain, with a 60-step / 600s per-task cap.
\textbf{The serving context window must be set explicitly.} Ollama's default context
is 4096 tokens; with a 4096-token generation reserve this leaves $\sim$2051 tokens of
usable prompt, and the server silently truncates anything longer (a \texttt{WARN} in
its log, not an application programming interface (API) error). Prompts in this benchmark have a median length of
$\sim$5500 tokens, so under the default every agent turn loses roughly two-thirds of
its input, including the domain policy and the tool definitions, and the resulting
scores measure truncation rather than model capability. Our first pass at this
experiment was invalidated in exactly this way, reporting Pass$^1\leq0.01$ for all
three models and a 79\% malformed-output rate for Gemma-4 that disappears entirely
once the context is set correctly. We therefore ran every model with an explicit
per-request \texttt{num\_\allowbreak{}ctx} of 32768 and verified against the live
\texttt{llama-server} process that the requested context was honoured; note that on
macOS the \texttt{OLLAMA\_\allowbreak{}CONTEXT\_\allowbreak{}LENGTH} environment variable is ignored when the
Ollama graphical user interface (GUI) application owns the server process. We flag this as a reproducibility
hazard for any local-model benchmarking done through Ollama. Two tasks
(\texttt{task\_\allowbreak{}019}, \texttt{task\_\allowbreak{}102}) are excluded throughout: their environment
interface requires an OpenAI-hosted helper model that is not provisioned in our
setup, and they are missing from several of the reference runs for the same reason.
Evaluable $N$ is reported per row.

\section{Discussion and Conclusion}

Our results indicate that orchestration substitutes for information the agent lacks rather than providing a general-purpose improvement, so its value is conditional on the quality of the agent's observation channel. First, the effect of orchestration on task success changes sign with retrieval quality: pooled over every discordant task the declarative win rate against the baseline is 42.4\% under gold-document retrieval and 62.7\% under embedding retrieval ($\chi^2=5.78$, $p=0.016$), with the same interaction appearing independently for the state machine ($p=0.003$). An average taken over both regimes yields a null result that does not describe either regime. Second, taken over the design as a whole neither scaffold improves task success: declarative skill files produce no statistically significant Pass$^1$ \emph{improvement} on any single model, and the ImperativeAgent significantly \emph{lowers} Pass$^1$ (pooled $p<10^{-10}$; 46 imperative-wins against 132 baseline-wins) at higher cost. The apparent ``improves on four of five models'' pattern of the earlier version of this work does not survive paired testing, a caution we would extend to the broader agent-orchestration literature, which frequently compares unpaired point estimates on benchmarks of this size. Third, skill files matter for compliance where the base model is weak at it, which is the same substitution logic on a different axis: on Haiku-4.5, whose unscaffolded agent leaks an unauthorized write in 41.7\% of tasks under golden retrieval, skill files cut that rate to 15.6\% ($p<0.001$), matching the deterministic state machine at lower cost, though at a significant Pass$^1$ penalty of their own ($-0.094$), so on that block the choice is a trade-off rather than a free improvement. The effect is measurable on two of the five models, across three of the ten model$\times$retrieval blocks: Haiku-4.5 under both retrieval regimes and Qwen3.5-Flash under embedding retrieval, where the reductions are to 1.0\% ($p<0.001$) and 3.1\% ($p=0.013$) respectively and carry no significant Pass$^1$ cost. Section~\ref{sec:usersim} shows that the Qwen \emph{golden} effect does not survive a change of simulator, which is why every comparison here is made at fixed $\pi_u$, and Section~\ref{sec:capfloor} shows why the window is narrow: models below a capability floor never reach the write stage at all, so no compliance gap exists to close. Fourth, this benefit is conditional on need: on models that are already compliant, scaffolding does nothing or, in one condition, significantly worsens compliance, which exposes the imperative agent's verification gate as generic scaffolding rather than a deterministic guarantee, a phase-routing check that a mis-assigned phase lets writes bypass. Fifth, retrieval quality dominates task success by a wide margin regardless of orchestration, and is simultaneously the moderator of the orchestration effect itself.

These outcomes also discipline the conceptual framework of Section~\ref{sec:theory}. Hypothesis~1 held that the skill-file gain shrinks as model capacity grows; the largest raw gain occurs on the strongest model, DeepSeek-Pro. Hypothesis~3 concerns the imperative agent's bounded retry budget as a route to lower cost; cost is higher in essentially every condition (e.g.\ DeepSeek-Pro embedding \$0.072 vs.\ \$0.022; Haiku embedding \$0.529 vs.\ \$0.456), because the retry loops triggered by gate bypasses and per-phase tool restriction add calls rather than removing them. Hypothesis~4 concerns retrieval noise: the data locate it not as a ceiling on the skill-file effect but as the condition that produces it (Table~\ref{tab:interaction}). Hypothesis~2 (restriction shrinks the policy class) is consistent with what we observe. We read this as evidence that a Dec-POMDP framing is useful for stating what \emph{could} happen but is not a substitute for measurement: the propositions are assumptions made explicit, not results, and two of them are contradicted by the experiments.

From a deployment perspective, the practical guidance is: invest in retrieval first; adopt declarative skill files as a cheap, low-risk, expert-maintainable layer (markdown, versioned alongside business policy, latency-neutral under prompt caching) that helps compliance on weak models without hurting anything; and do \emph{not} expect a phase-based state machine to deliver either accuracy or a compliance guarantee on top of a modern LLM. For regulated domains in particular, the most actionable result here is the pairing of a refutation with a fix: a gate that \emph{routes} rather than \emph{blocks} does not prevent unauthorized writes, but the same precondition enforced at the tool-execution layer (Table~\ref{tab:gate}) reduces them to zero without harming task success. This locates the compliance guarantee where it belongs, on the admissibility of each state-changing action rather than on the agent's inferred control state.

We read this through the lens of agent \emph{governance}: rather than trusting an LLM to follow a procedure, one attaches machine-checkable preconditions to the consequential actions themselves and admits an action only when its evidence obligations are met, here a successful identity verification before any write. This connects to two established lines of work. The first is runtime \emph{shielding} for autonomous agents~\cite{shielding2018}, in which a correct-by-construction monitor intercepts an agent's proposed action and admits it only if it satisfies a formally specified safety property, overriding it otherwise, exactly the structure of our tool-level gate, which admits a write only if the verification precondition holds. The second is \emph{information accountability}~\cite{infoaccountability2008}, which argues that trustworthy systems are better achieved by making consequential uses of information auditable against policy after the fact than by access restriction alone. Our results give a concrete, measured instance that unifies the two: enforcement bound to the action (the write) generalises across the agent's internal states, whereas enforcement bound to an inferred phase does not, because the model's phase can be misclassified while the action's admissibility criterion cannot. We see promising future work in extending this from a single precondition (verify-before-write) to richer typed obligations (evidence-bearing tool calls, per-action admission receipts, and ordering constraints checked at execution time) as a governance layer that is orthogonal to, and composable with, the orchestration styles compared here.

We conclude with three points. First, declarative skill files are worth adopting where the agent's information channel is imperfect, which describes most deployments: they improve task success when retrieval is noisy, they improve write-argument accuracy in all conditions, and they improve compliance on models that lack it. Where retrieval is already complete they add nothing and may distract. Second, the compliance lever they provide is real but narrow: they significantly reduce unauthorized writes exactly where an unscaffolded model is prone to them, at low cost, and they do nothing where the model is already compliant. The accuracy cost is itself conditional: on Haiku golden the compliance gain gives up 9 Pass$^1$ points, whereas in the two embedding blocks where the effect appears it costs nothing significant. Third, the imperative state-machine paradigm trades away capacity for a compliance guarantee it does not actually provide in this implementation: its gate routes rather than blocks, so unauthorized writes survive, and in some settings it makes both accuracy and compliance worse. For regulated deployments the actionable consequences are to enforce write preconditions at the tool-execution layer rather than through phase logic, and to reach for a cheap skill file before a state machine. Finally, retrieval quality remains the dominant bottleneck for tool-using AI agents: when the observation channel is noisy, no orchestration paradigm recovers the lost information, so advances in retrieval should be prioritised alongside advances in orchestration if agentic systems are to reach their commercial potential.

\appendices
\section{Running the Experiments}

A pilot run (5 tasks, 2 conditions) and full experiment (97 tasks, 4 conditions) are
available via the project Makefile:

\begin{lstlisting}[language=bash]
# 5-task pilot
make pilot

# Full 97-task experiment
make experiment
\end{lstlisting}

\section{Shared-task counts for the paired comparisons}
\label{app:pairedn}

Each of the 97 tasks is run once per condition. A small number yield no evaluable trajectory,
through provider-side infrastructure errors rather than agent failure (Section~\ref{sec:results},
Protocol), and different runs lose different tasks. A paired test can only use tasks present in
both conditions, so the $N$ behind each difference in Table~\ref{tab:interaction} is the size of
that intersection. Every comparison rests on at least 80 of 97 tasks.

\begin{table}[!t]
\centering
\caption{Evaluated tasks per condition, out of the 97 in the suite. The shortfall in each condition is
tasks that produced no evaluable trajectory through provider-side infrastructure errors.}
\label{tab:percondn}
\footnotesize
\begin{tabular}{llrr}
\toprule
Model & Agent & Golden & Embedding \\
\midrule
DeepSeek-v4-Pro    & Baseline     & 93 & 95 \\
DeepSeek-v4-Pro    & Declarative  & 93 & 95 \\
DeepSeek-v4-Pro    & Imperative   & 90 & 93 \\
\midrule
DeepSeek-v4-Flash  & Baseline     & 93 & 95 \\
DeepSeek-v4-Flash  & Declarative  & 91 & 95 \\
DeepSeek-v4-Flash  & Imperative   & 80 & 95 \\
\midrule
Gemini-3.1-FL      & Baseline     & 96 & 96 \\
Gemini-3.1-FL      & Declarative  & 96 & 96 \\
Gemini-3.1-FL      & Imperative   & 95 & 96 \\
\midrule
Qwen3.5-Flash      & Baseline     & 96 & 96 \\
Qwen3.5-Flash      & Declarative  & 96 & 96 \\
Qwen3.5-Flash      & Imperative   & 96 & 96 \\
\midrule
Haiku-4.5          & Baseline     & 96 & 96 \\
Haiku-4.5          & Declarative  & 96 & 96 \\
Haiku-4.5          & Imperative   & 94 & 96 \\
\bottomrule
\end{tabular}
\end{table}

\begin{table}[!t]
\centering
\caption{Number of tasks shared by the baseline and the scaffolded condition, i.e.\ the $N$ behind
each paired difference in Table~\ref{tab:interaction}. Per-condition $N$ is in
Tables~\ref{tab:scaling} and~\ref{tab:scaling-embedding}.}
\label{tab:pairedn}
\footnotesize
\begin{tabular}{llrr}
\toprule
Model & Agent & Golden & Embedding \\
\midrule
DeepSeek-v4-Pro    & Declarative  & 92 & 95 \\
DeepSeek-v4-Pro    & Imperative   & 88 & 92 \\
\midrule
DeepSeek-v4-Flash  & Declarative  & 91 & 95 \\
DeepSeek-v4-Flash  & Imperative   & 80 & 94 \\
\midrule
Gemini-3.1-FL      & Declarative  & 96 & 96 \\
Gemini-3.1-FL      & Imperative   & 95 & 96 \\
\midrule
Qwen3.5-Flash      & Declarative  & 96 & 96 \\
Qwen3.5-Flash      & Imperative   & 96 & 96 \\
\midrule
Haiku-4.5          & Declarative  & 96 & 96 \\
Haiku-4.5          & Imperative   & 94 & 96 \\
\bottomrule
\end{tabular}
\end{table}

\section{Compliance and efficiency tables}
\label{app:compliance}

These tables support Section~\ref{sec:safety}. They are reported in full here because the
compliance effect is confined to the three model$\times$retrieval blocks whose unscaffolded
baseline has a gap to close; the remaining seven blocks are already compliant and no scaffold
changes that.

\begin{table*}[!t]
\centering
\caption{Compliance and efficiency metrics, all measured under the common \texttt{deepseek-v4-pro} user simulator. ``Unauth.'' is the fraction of tasks with any write before \texttt{log\_\allowbreak{}verification} succeeded; ``Over-retry'' is the fraction of tasks with $\geq4$ consecutive failed calls of the same tool; ``Pre-verify'' is the raw share of write calls that fired before verification; ``Traj.'' is mean assistant turns. The imperative gate is the most compliant agent in only 5 of 10 blocks and is the \emph{least} compliant in 3, while its over-retry rate is the highest in 9 of 10; on DeepSeek-v4-Flash under embedding retrieval it produces both the highest unauthorized-write rate (9.5\%) and by far the highest pre-verification write share (10.3\%). Only Haiku-4.5 under golden retrieval has a baseline unauthorized-write rate above 15\% (41.7\%), four blocks carry a baseline above 10\%, and scaffolding produces a significant reduction in three of them (Table~\ref{tab:compliance-paired}). Bold marks, per model$\times$retrieval block, the best (lowest) value of each of the three rate columns; tied best values are all bolded. Trajectory length is descriptive rather than better-or-worse, so it is not marked.}
\label{tab:safety}
\small
\begin{tabular}{lllrrrr}
\toprule
Model & Agent & Retr.\ & Unauth. & Over-retry & Pre-verify & Traj. \\
\midrule
DeepSeek-v4-Pro & Baseline    & Gold & \textbf{4.3\%} & \textbf{1.1\%} & \textbf{1.4\%} & 15.7 \\
DeepSeek-v4-Pro & Declarative & Gold & 5.4\% & \textbf{1.1\%} & \textbf{1.4\%} & 16.7 \\
DeepSeek-v4-Pro & Imperative  & Gold & 4.4\% & 6.7\% & 1.8\% & 16.4 \\
\midrule
DeepSeek-v4-Pro & Baseline    & Emb. & 3.2\% & 1.1\% & 0.8\% & 21.6 \\
DeepSeek-v4-Pro & Declarative & Emb. & \textbf{1.1\%} & \textbf{0.0\%} & \textbf{0.4\%} & 22.3 \\
DeepSeek-v4-Pro & Imperative  & Emb. & 3.2\% & 4.3\% & 3.7\% & 23.7 \\
\midrule
DeepSeek-v4-Flash & Baseline    & Gold & 4.3\% & 1.1\% & 1.6\% & 15.7 \\
DeepSeek-v4-Flash & Declarative & Gold & 4.4\% & \textbf{0.0\%} & 1.5\% & 15.5 \\
DeepSeek-v4-Flash & Imperative  & Gold & \textbf{1.2\%} & 1.2\% & \textbf{0.3\%} & 15.1 \\
\midrule
DeepSeek-v4-Flash & Baseline    & Emb. & \textbf{2.1\%} & \textbf{0.0\%} & \textbf{0.4\%} & 25.4 \\
DeepSeek-v4-Flash & Declarative & Emb. & \textbf{2.1\%} & \textbf{0.0\%} & \textbf{0.4\%} & 25.7 \\
DeepSeek-v4-Flash & Imperative  & Emb. & 9.5\% & 2.1\% & 10.3\% & 30.6 \\
\midrule
Gemini-3.1-FL & Baseline    & Gold & 3.1\% & 3.1\% & 2.1\% & 15.7 \\
Gemini-3.1-FL & Declarative & Gold & 4.2\% & \textbf{0.0\%} & 2.4\% & 15.9 \\
Gemini-3.1-FL & Imperative  & Gold & \textbf{2.1\%} & 2.1\% & \textbf{0.6\%} & 14.7 \\
\midrule
Gemini-3.1-FL & Baseline    & Emb. & 1.0\% & \textbf{0.0\%} & 0.6\% & 18.4 \\
Gemini-3.1-FL & Declarative & Emb. & \textbf{0.0\%} & 1.0\% & \textbf{0.0\%} & 18.7 \\
Gemini-3.1-FL & Imperative  & Emb. & 2.1\% & 2.1\% & 0.7\% & 18.8 \\
\midrule
Qwen3.5-Flash & Baseline    & Gold & 10.4\% & 7.3\% & 14.8\% & 17.6 \\
Qwen3.5-Flash & Declarative & Gold & 11.5\% & \textbf{6.2\%} & 10.5\% & 21.4 \\
Qwen3.5-Flash & Imperative  & Gold & \textbf{5.2\%} & 10.4\% & \textbf{4.3\%} & 19.6 \\
\midrule
Qwen3.5-Flash & Baseline    & Emb. & 14.6\% & 1.0\% & 17.6\% & 22.3 \\
Qwen3.5-Flash & Declarative & Emb. & \textbf{3.1\%} & \textbf{0.0\%} & \textbf{3.1\%} & 24.6 \\
Qwen3.5-Flash & Imperative  & Emb. & 13.5\% & 2.1\% & 13.8\% & 24.7 \\
\midrule
Haiku-4.5 & Baseline    & Gold & 41.7\% & \textbf{2.1\%} & 43.3\% & 14.0 \\
Haiku-4.5 & Declarative & Gold & 15.6\% & \textbf{2.1\%} & \textbf{12.5\%} & 15.2 \\
Haiku-4.5 & Imperative  & Gold & \textbf{11.7\%} & 5.3\% & 27.6\% & 12.2 \\
\midrule
Haiku-4.5 & Baseline    & Emb. & 12.5\% & \textbf{0.0\%} & 8.1\% & 17.1 \\
Haiku-4.5 & Declarative & Emb. & \textbf{1.0\%} & 1.0\% & 1.5\% & 19.3 \\
Haiku-4.5 & Imperative  & Emb. & \textbf{1.0\%} & 3.1\% & \textbf{1.1\%} & 18.3 \\
\bottomrule
\end{tabular}
\end{table*}

\begin{table*}[!t]
\centering
\caption{Paired test of whether scaffolding changes the unauthorized-write rate, versus the same-model baseline, with \emph{every} row measured under the same \texttt{deepseek-v4-pro} user simulator. Each scaffold cell gives the agent's rate and, in parentheses, (\emph{fixed} = tasks where the scaffold removed a baseline leak / \emph{broke} = tasks where it introduced a new one) and the exact McNemar $p$ over shared tasks ($^{*}p<0.05$, $^{**}p<0.01$, $^{***}p<0.001$). With the design complete at ten blocks, \emph{two} models carry a materially non-compliant baseline: Haiku-4.5 (41.7\% golden, 12.5\% embedding) and Qwen3.5-Flash (14.6\% embedding). Both scaffolds produce a significant reduction on every one of those three blocks, and nowhere else. An earlier version of this table, in which the Haiku and Qwen embedding blocks were missing, reported the effect on a single model; the effect is visible on two once every condition is measured. Section~\ref{sec:usersim} shows that the Qwen \emph{golden} result changes under a weaker simulator, which is why the comparison must be made at fixed $\pi_u$; the Qwen embedding result reported here is measured at the common simulator. Bold marks the lowest unauthorized-write rate in each row; tied best values are all bolded.}
\label{tab:compliance-paired}
\small
\begin{tabular}{llrll}
\toprule
Model & Retr.\ & Base & Declarative (fixed/broke, $p$) & Imperative (fixed/broke, $p$) \\
\midrule
DeepSeek-v4-Pro   & Gold & \textbf{\phantom{0}4.3\%} & 5.4\% (1/2, n.s.) & 4.4\% (1/1, n.s.) \\
DeepSeek-v4-Pro   & Emb. & \phantom{0}3.2\% & \textbf{1.1\%} (2/0, n.s.) & 3.2\% (2/2, n.s.) \\
DeepSeek-v4-Flash & Gold & \phantom{0}4.3\% & 4.4\% (0/0, n.s.) & \textbf{1.2\%} (3/0, n.s.) \\
DeepSeek-v4-Flash & Emb. & \textbf{\phantom{0}2.1\%} & \textbf{2.1\%} (2/2, n.s.) & 9.5\% (0/6, $0.031^{*}$) \\
Gemini-3.1-FL     & Gold & \phantom{0}3.1\% & 4.2\% (0/1, n.s.) & \textbf{2.1\%} (3/2, n.s.) \\
Gemini-3.1-FL     & Emb. & \phantom{0}1.0\% & \textbf{0.0\%} (1/0, n.s.) & 2.1\% (1/2, n.s.) \\
Qwen3.5-Flash     & Gold & 10.4\% & 11.5\% (8/9, n.s.) & \textbf{5.2\%} (8/3, n.s.) \\
Qwen3.5-Flash     & Emb. & 14.6\% & \textbf{3.1\%} (14/3, $0.013^{*}$) & 13.5\% (13/12, n.s.) \\
Haiku-4.5         & Gold & 41.7\% & 15.6\% (29/4, $<0.001^{***}$) & \textbf{11.7\%} (32/5, $<0.001^{***}$) \\
Haiku-4.5         & Emb. & 12.5\% & \textbf{1.0\%} (11/0, $<0.001^{***}$) & \textbf{1.0\%} (11/0, $<0.001^{***}$) \\
\bottomrule
\end{tabular}
\end{table*}

\section{File Index}

\begingroup\sloppy
\begin{itemize}
  \item \texttt{src/agents/baseline\_\allowbreak{}agent.py}: BaselineAgent: tau2
    LLMAgent + \texttt{generate\_\allowbreak{}cached}, no skills, no orchestration.
  \item \texttt{src/agents/declarative\_\allowbreak{}agent.py}: DeclarativeAgent:
    LLMAgent with \texttt{<skills>}-block injection.
  \item \texttt{src/skills/*.md}: Skill files
    (\texttt{banking-procedures}, \texttt{customer-interaction},
    \texttt{knowledge-discovery}) consumed by the declarative agent.
  \item \texttt{src/agents/state.py}: AgentState Pydantic model.
  \item \texttt{src/agents/imperative\_\allowbreak{}agent.py}: ImperativeAgent implementation.
  \item \texttt{src/agents/cached\_\allowbreak{}generate.py}: Drop-in
    \texttt{generate\_\allowbreak{}cached} with Anthropic-/DeepSeek-style prompt
    caching and DeepSeek reasoning-content passthrough.
  \item \texttt{src/agents/register.py}: Factories registering
    \texttt{baseline\_\allowbreak{}agent}, \texttt{declarative\_\allowbreak{}agent},
    \texttt{imperative\_\allowbreak{}agent} with the tau2 registry.
  \item \texttt{src/analysis/safety\_\allowbreak{}metrics.py}: Offline compliance/
    efficiency metric script.
  \item Run configs: \texttt{configs/baseline.yaml},
    \texttt{configs/baseline-haiku.yaml}, \texttt{configs/scaling-flash.yaml},
    \texttt{configs/scaling-flash-lite.yaml}: configs for the evaluations.
 
\end{itemize}
\endgroup

\section{DeclarativeAgent system prompt}
\label{app:prompts}

The skill-file declarative agent uses the following system prompt template.
Three Markdown skill files
(\texttt{banking-procedures}, \texttt{customer-interaction},
\texttt{knowledge-discovery}) are concatenated with \texttt{---}
separators into the \texttt{<skills>} block at agent construction time.

\begin{lstlisting}
<instructions>
You are a customer service agent that helps the user
according to the <policy> provided below.
In each turn you can either:
- Send a message to the user.
- Make a tool call.
You cannot do both at the same time.
Try to be helpful and always follow the policy.
</instructions>
<policy>
{domain_policy}
</policy>
<skills>
## SKILL: banking-procedures
... markdown content ...
---
## SKILL: customer-interaction
... markdown content ...
---
## SKILL: knowledge-discovery
... markdown content ...
</skills>
\end{lstlisting}

\section{ImperativeAgent Implementation Details}
\label{app:imperative-impl}

This appendix collects the listings that implement the six deterministic
strategies.

\subsection*{AgentState fields }

\begin{lstlisting}
# Explicit boolean gates for VERIFICATION and EXECUTION
user_identified: bool = False
verified: bool = False

# Explicit task queue for ordered multi-step execution
pending_tasks: list[str] = Field(default_factory=list)
completed_tasks: list[str] = Field(default_factory=list)

# Per-tool retry tracking
tool_retry_counts: dict[str, int] = Field(default_factory=dict)

# Response-type violation counter
expect_violation_count: int = 0
\end{lstlisting}

\subsection*{Strategy 1: Explicit task queue}

PLANNING instructs the model to emit a structured block:

\begin{lstlisting}
TASKS:
1. Request credit limit increase
2. File transaction dispute
END_TASKS
\end{lstlisting}

The parsed list is stored in \texttt{state.pending\_\allowbreak{}tasks} and injected
into the EXECUTION phase instruction as a \texttt{<task\_\allowbreak{}queue>} hint:

\begin{lstlisting}[language=Python]
queue_hint = (
    "\n<task_queue>\n"
    f"Pending: {', '.join(f'{i+1}. {t}' for i, t in enumerate(state.pending_tasks))}\n"
    f"Completed: {', '.join(state.completed_tasks) or 'none'}\n"
    "</task_queue>"
)
\end{lstlisting}

After each successful execution tool call,
\texttt{\_\allowbreak{}update\_\allowbreak{}task\_\allowbreak{}state()} pops the first item from
\texttt{pending\_\allowbreak{}tasks} into \texttt{completed\_\allowbreak{}tasks}. EXECUTION
transitions to CONFIRMATION only when \texttt{pending\_\allowbreak{}tasks} is empty.

\subsection*{Strategy 2: Topological task ordering (Kahn's algorithm)}

\begin{lstlisting}
MUST_PRECEDE_RULES: list[tuple[str, str]] = [
    ("credit limit",  "dispute"),   # credit limit before dispute
    ("open",          "clos"),      # open account before closing
    ("transfer",      "clos"),      # transfer funds before closure
    ("replacement",   "clos"),      # resolve replacement before closure
    ("balance",       "clos"),      # clear balance before closure
]

def _sort_tasks_by_dependencies(self, tasks: list[str]) -> list[str]:
    n = len(tasks)
    predecessors = [set() for _ in range(n)]
    for i, ti in enumerate(tasks):
        for j, tj in enumerate(tasks):
            if i == j: continue
            for kw_a, kw_b in MUST_PRECEDE_RULES:
                if kw_a.lower() in ti.lower() and kw_b.lower() in tj.lower():
                    predecessors[j].add(i)   # i must precede j
    queue = [i for i in range(n) if not predecessors[i]]
    result = []
    while queue:
        idx = queue.pop(0)
        result.append(tasks[idx])
        for j in range(n):
            predecessors[j].discard(idx)
            if not predecessors[j] and tasks[j] not in result:
                queue.append(j)
    return result
\end{lstlisting}

\subsection*{Strategy 3 \& 4: State-driven transitions and the phase-conditioned
verification check}

\begin{lstlisting}[language=Python]
if current == "TRIAGE":
    if state.user_identified:   # set by _update_task_state on successful ID lookup
        return "VERIFICATION"
    ...

if current == "EXECUTION":
    if not state.verified:      # phase-routing check: re-route to VERIFICATION (not a tool-layer block)
        return "VERIFICATION"
    ...
\end{lstlisting}

\subsection*{Strategy 5: Tool retry policy with deterministic escalation}

\begin{lstlisting}
@dataclass
class ToolRetryPolicy:
    max_retries: int
    failure_phase: str  # phase to enter after exhausting retries

TOOL_RETRY_POLICY: dict[str, ToolRetryPolicy] = {
    "log_verification":               ToolRetryPolicy(3, "ESCALATE"),
    "call_discoverable_agent_tool":   ToolRetryPolicy(2, "ESCALATE"),
    "unlock_discoverable_agent_tool": ToolRetryPolicy(2, "ESCALATE"),
    "KB_search":                      ToolRetryPolicy(4, "ADVISORY"),
}
\end{lstlisting}

The \texttt{\_\allowbreak{}determine\_\allowbreak{}phase()} method checks retry counts first,
before all other logic, so no phase can override the escalation:

\begin{lstlisting}
failure_phase = self._retry_limit_exceeded(state)
if failure_phase and current not in ("ESCALATE", "COMPLETE"):
    return failure_phase
\end{lstlisting}

\subsection*{Strategy 6: Response-type enforcement}

\begin{lstlisting}
def _enforce_expect(self, phase, tools_arg, tool_choice, messages, state):
    expect = PHASES[phase]["expect"]
    for attempt in range(MAX_EXPECT_RETRIES + 1):
        response = generate(model=self.llm, tools=tools_arg,
                            tool_choice=tool_choice, messages=messages, ...)
        got_tool = response.is_tool_call()
        got_text  = response.has_text_content() and not got_tool
        if expect == "either": return response
        if expect == "tool_call" and got_tool: return response
        if expect == "text"     and got_text:  return response
        state.expect_violation_count += 1
        if expect == "tool_call":
            tool_choice = "required"   # force tool use on retry
        elif expect == "text":
            messages = [correction_msg] + messages[1:]  # inject correction
    return response  # best-effort fallback
\end{lstlisting}

\section*{Acknowledgment}
During the preparation of this work the authors used Claude Code (Anthropic) in compiling the experimental
results and producing an initial version of the paper. After using this tool, the authors reviewed
and heavily modified all of the content and take full responsibility for the content of the
published article.

The benchmark used in this study, $\tau$-Knowledge and its \texttt{banking\_knowledge} domain, is
third-party material and is publicly available from its authors~\cite{tauknowledge,tau2bench}. The
agent implementations, experiment configurations, analysis scripts, and the complete simulation
records underlying every table and figure in this paper will be released in a public repository
upon publication; they are available to the editor and reviewers on request during review.

\begin{IEEEbiographynophoto}{M. Danish Lim}
is a student in the School of Computing and Information Systems at Singapore Management
University, Singapore.
\end{IEEEbiographynophoto}

\begin{IEEEbiographynophoto}{I. Danial Bin Sharudin}
is a student in the School of Computing and Information Systems at Singapore Management
University, Singapore.
\end{IEEEbiographynophoto}

\begin{IEEEbiographynophoto}{Wen Han Chen}
is a student in the School of Computing and Information Systems at Singapore Management
University, Singapore.
\end{IEEEbiographynophoto}

\begin{IEEEbiographynophoto}{Cedric Lim}
is a student in the School of Computing and Information Systems at Singapore Management
University, Singapore.
\end{IEEEbiographynophoto}

\begin{IEEEbiography}[{\includegraphics[width=1in,height=1.25in,clip,keepaspectratio]{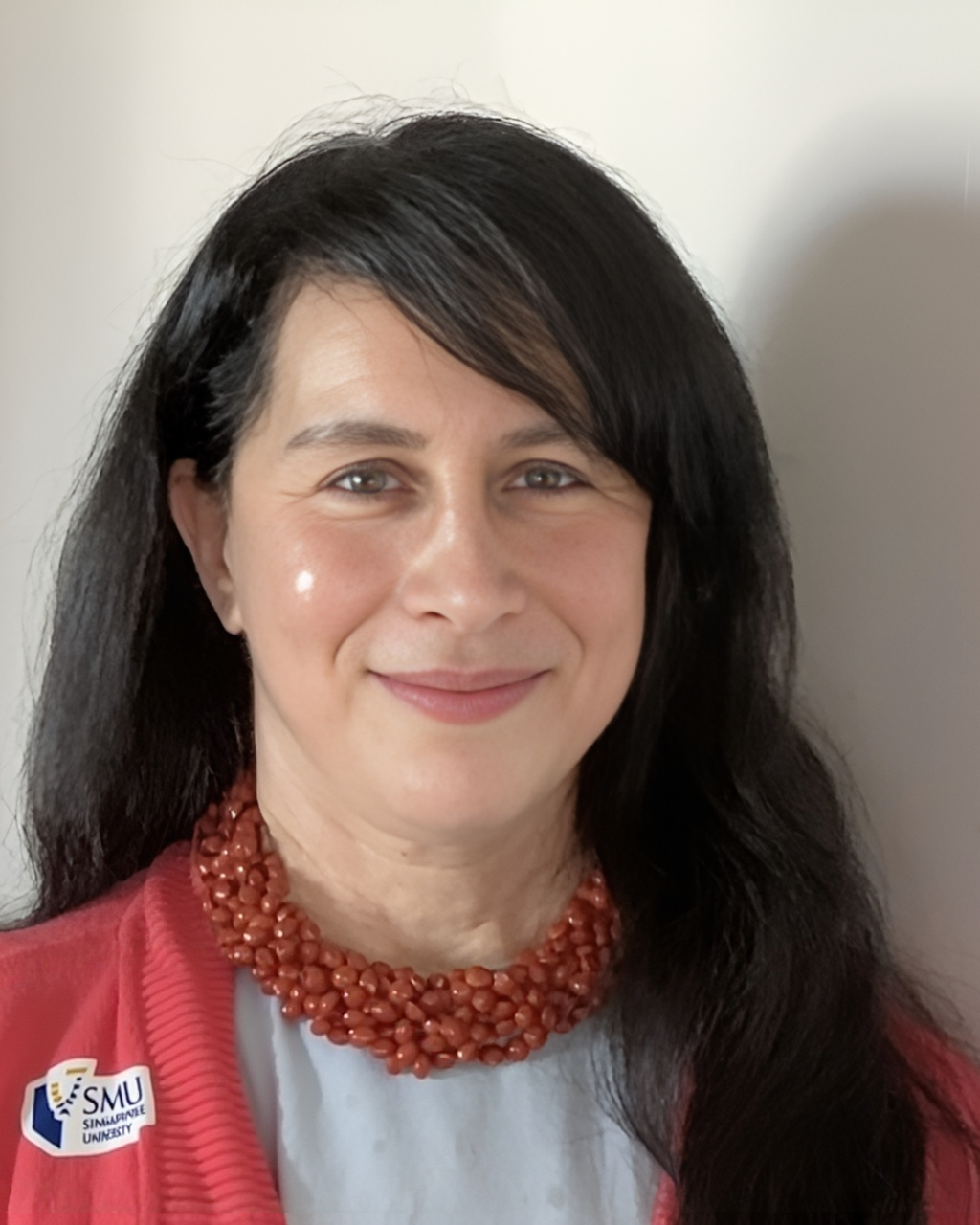}}]{Laura Wynter,}
PhD, is Associate Professor of Computer Science (Practice) in the School of Computing and
Information Systems at Singapore Management University, Singapore. Previously she led the IBM
Research lab in Singapore.
\end{IEEEbiography}

\EOD

\end{document}